\documentclass[10pt,onecolumn,letterpaper]{article}

\usepackage{cvpr}
\usepackage{times}
\usepackage{epsfig}
\usepackage{graphicx}
\usepackage{amsmath}
\usepackage{amssymb}
\usepackage[title]{appendix}
\usepackage{placeins} 
\usepackage{dblfloatfix} 
\usepackage{subcaption} 
\usepackage{enumitem} 

\newcommand\figref{Figure~\ref}

\usepackage[pagebackref=true,breaklinks=true,letterpaper=true,colorlinks,bookmarks=false]{hyperref}

\cvprfinalcopy 

\def\cvprPaperID{****} 

\ifcvprfinal\fi
\begin{document}

\title{Saccade Sequence Prediction: Beyond Static Saliency Maps}

\author{Calden Wloka\\
Department of Electrical Engineering and Computer Science\\
York University, Toronto, Canada\\
{\tt\small calden@cse.yorku.ca}
\and
Iuliia Kotseruba\\
Department of Electrical Engineering and Computer Science\\
York University, Toronto, Canada\\
{\tt\small yulia\_k@cse.yorku.ca}\\
\and
John K. Tsotsos\\
Department of Electrical Engineering and Computer Science\\
York University, Toronto, Canada\\
{\tt\small tsotsos@cse.yorku.ca}
}

\maketitle

\begin{abstract}
Visual attention is a field with a considerable history, with eye movement control and prediction forming an important subfield. Fixation modeling in the past decades has been largely dominated computationally by a number of highly influential bottom-up saliency models, such as the Itti-Koch-Niebur model. The accuracy of such models has dramatically increased recently due to deep learning. However, on static images the emphasis of these models has largely been based on non-ordered prediction of fixations through a saliency map. Very few implemented models can generate temporally ordered human-like sequences of saccades beyond an initial fixation point. Towards addressing these shortcomings we present STAR-FC, a novel multi-saccade generator based on a central/peripheral integration of deep learning-based saliency and lower-level feature-based saliency. We have evaluated our model using the CAT2000 database, successfully predicting human patterns of fixation with equivalent accuracy and quality compared to what can be achieved by using one human sequence to predict another. This is a significant improvement over fixation sequences predicted by state-of-the-art saliency algorithms.
\end{abstract}

\section{Introduction}

\label{sec:Intro}
Most applications in computer vision function primarily in a passive way; algorithms are applied to static images or pre-recorded video sequences without control over what visual data is acquired next. However, it has long been recognized that eye movements are an integral aspect to human vision \cite{Kowler2011}, with diverse functionality ranging from the enhanced extraction of features via microsaccadic motion \cite{KuangEtAl2012} through high-level strategies for optimal information gathering \cite{NajemnikGeisler2005}. It is this latter aspect which is of particular interest to the field of computer vision; active control over the acquisition of image data is fundamental to efficiently developing more robust and general computer vision solutions for unconstrained environments \cite{Tsotsos1992,BajcsyEtAl2016}.

\begin{figure}[!htbp]
	\begin{center}
		\includegraphics[width=0.9\linewidth]{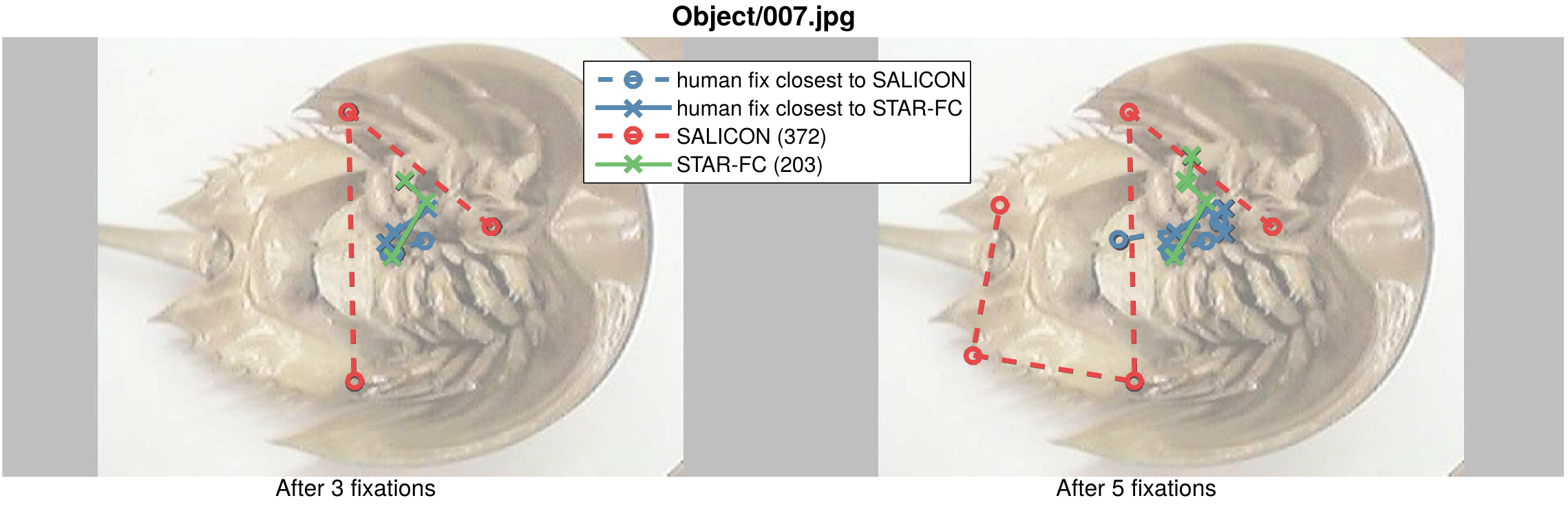} 
	\end{center}
	\vspace{-3mm}
	\caption{An example image from the CAT2000 dataset with overlaid fixation sequences. We show sequences predicted by our STAR-FC model (green) and SALICON (red), along with the human sequence which best matches STAR-FC (solid blue with x's) and which best matches SALICON (dashed blue with o's). Euclidean distances between each model and its closest human sequences after five fixations are noted in parentheses of in the legend.
	 \label{fig:CrabFixExample}}
	\vspace{-4mm}
\end{figure}

Our work presented here develops and extends the Selective Tuning Attentive Reference model Fixation Controller (STAR-FC): an explicit model of human saccadic control \cite{TsotsosEtAl2016}. In order to more easily compare with prior efforts in fixation prediction, we concentrate on the free-viewing paradigm, but nevertheless specify our control network in a manner which provides explicit extensibility for task-based tuning and top-down attentional control. By providing an extensive model of human fixation control which includes a number of aspects normally neglected by the saliency literature, including an explicit transform to account for anisotropic retinal acuity, we are able to produce explicit fixation sequences with greater fidelity to those of humans than is seen by traditional saliency approaches (see Figure \ref{fig:CrabFixExample} for an example, or Appendix \ref{sec:ExamplesApdx} for more). Altogether, we offer the following contributions:
\begin{itemize}
{
	\item{A novel computational fixation model which outperforms traditional saliency map models for explicit sequences prediction.}
	\item{A descriptive model of fixation control which may be used to better explore the function of early human attention.}
	\item{A flexible design which may be parametrically tuned to match the specific experimental conditions under which eye tracking data is obtained.}
}
\end{itemize}

\subsection{Background}
\label{sub:Background}
Eye movement control and early visual attention have frequently been conflated, particularly within the computational saliency literature. The term "saliency map" was coined in \cite{KochUllman1985} in the context of covert, pre-attentive visual processing. Due to the significant challenge of obtaining a suitable source of ground-truth data with which to validate a map of pre-attentive processing, focus shifted to predicting fixation locations \cite{IttiKochNiebur1998}. Since then, many saliency algorithms have been proposed, ranging from information theoretic principles \cite{BruceTsotsos2007}, efficient coding \cite{GarciaEtAl2012}, spectral analysis \cite{HouKoch2012}, or processing pipelines driven largely by empirical performance \cite{RicheEtAl2012}, to name a few. One of the earliest machine learning efforts used a collection of low-, mid-, and high-level features as inputs to an SVM classifier in order to classify pixels as either salient or not \cite{JuddEtAl2009}. More recently, however, deep learning networks have come to dominate the field \cite{HuangEtAl2015,KummererEtAl2015,KruthiventiEtAl2015,LiuEtAl2015}. 

One schism which has formed within saliency research is whether the focus should be on locations or objects. Much of this split originated from the claim of Einh{\"a}user \etal \cite{EinhaeuserEtAl2008} that objects themselves actually predict fixations better than feature-based pixel saliency. This led to a number of approaches including those which seek to generate saliency maps based on class-generic object detectors (\eg \cite{AlexeEtAl2010} and subsequent extensions to saliency \cite{ChangEtAl2011}) and those which train and test saliency algorithms explicitly using object masks rather than fixation data (\eg, \cite{LiYu2016}). However, there has been push-back against this object-centric view, with Borji \etal \cite{BorjiEtAl2013} arguing that the original findings of Einh{\"a}user \etal were based largely on the metric used to measure performance. Given the focus of this paper on the explicit generation of saccade sequences, we test our algorithm performance against fixation data rather than object masks, but do take the view that there is a balance to be struck between pixel-level feature effects and higher-level object detection. This is discussed further in Section \ref{sub:Architecture}.

While our goal differs from the standard manner in which saliency algorithms are applied and evaluated, we compare performance against them in order to emphasize the importance of our novel perspective. Static saliency maps have previously been used to generate explicit fixation sequences, such as Itti and Koch's \cite{IttiKoch2000} search system which couples Winner-Take-All (WTA) selection to a simple inhibition of return scheme. The connection between explicit eye movement patterns and saliency maps was explored from a different direction by \cite{TatlerVincent2009}, in which a saliency algorithm independent of the visual input was based on statistical regularities in eye movements. Despite the lack of visual processing, it nevertheless demonstrated comparable or better performance than the Itti-Koch-Niebur (IKN) saliency model \cite{IttiKochNiebur1998}, suggesting that fixation location may be driven as much by the underlying motor control of the eye as it is by visual information.

Outside of the saliency literature there are a number of eye movement control models. However, such models are usually dedicated to a specific subset of eye movements (such as smooth pursuit \cite{PolaWyatt1991}, the optokinetic reflex \cite{DistlerHoffmann2011}, or 3D gaze shifts \cite{CrawfordKlier2011}) or neural component (such as the role of the superior colliculus \cite{WhiteMunoz2011}, cerebellum \cite{Miles1991} or the basal ganglia \cite{VokounEtAl2011}) without a clear path of extension or inclusion of attentional control. Tsotsos \etal \cite{TsotsosEtAl2016} provide a more general formulation of attentional control with a focus on saccadic sequences. Nevertheless, the implementation of their model provides only a largely qualitative demonstration of efficacy over a single image. We build upon the theoretical formulation laid out by \cite{TsotsosEtAl2016}, extending the architecture to function over a broad range of natural images which allows for a quantitative analysis of performance. See Section \ref{sub:Architecture} for a more thorough description of our model.

\subsection{Applications of Fixation Prediction}
\label{sub:Applications}
Early interest in saccadic sequences was heavily influenced by Noton and Stark's \emph{scanpath theory} \cite{NotonStark1971}, which posited that the explicit spatiotemporal structure of eye movements drove memory encoding for visual patterns and subsequent retrieval. However, challenges to this view have arisen over the years, with experimental evidence showing that there is no recognition advantage conferred by the use of one's own fixation locations versus those of another viewer nor by the retention of the temporal order of fixation \cite{FoulshamKingstone2013}. These results certainly support the traditional approach to saliency evaluation which predominantly seeks to evaluate algorithms on prediction effectiveness over a static ground-truth fixation cloud, disregarding individual source and temporal characteristics of the fixations.

However, scanpath theory was largely devoted to the memory encoding and recall of images. Even if visual memory is not heavily influenced by scanpaths, there are nevertheless a number of applications for which explicit fixation sequence modeling and prediction is very valuable. For example, motivated by the very short window of consumer attention to most advertisements, commercial applications of saliency analysis already include predicted sequences of the first several fixations \cite{VAS_Sample2015}, despite validation using only traditional ROC methods which do not measure the efficacy of sequence prediction \cite{VAS_Validation2010}. Understanding fixation locations has also gained recent interest in the area of science communication and policy making, particularly for graphical figures \cite{HaroldEtAl2016}. Even more so than in advertising, the sequence of fixations over a graphical figure becomes important for understanding whether and how viewers are understanding the information contained.

As previously mentioned, understanding the control of human eye movements may additionally be highly instructive in robotic visual systems with active camera control such as robotic search \cite{RasouliTsotsos2014}. This is particularly useful for applications with anisotropic sensors which could be considered analogous to the anisotropy present within the human retina, such as omnidirectional camera systems which introduce a high degree of spatial distortion unevenly across the visual field \cite{GasparEtAl2000} or a two-camera visual input system which combines high- and low-resolution streams to effectively maintain a wide field of view without sacrificing the ability to acquire high acuity detail over a targeted region \cite{ElderEtAl2006}. Furthermore, as robotic applications increase their focus on social interactions, it becomes important not only to accurately attend to relevant information during an interaction, but also to provide socially important cues through body language such as gaze location \cite{mavridis2007grounded}. Robotic modelling of joint attention has previously been improved through the application of saliency \cite{YuecelEtAl2013}, and can likely be further improved with a more complete gaze model. Accurate modelling of joint attention between parties has wide reaching ramifications, from self-driving vehicles \cite{KotserubaEtAl2016} to the handover of physical objects \cite{MoonEtAl2014}.

\section{Methods}
\label{sec:Methods}

\subsection{System Architecture}
\label{sub:Architecture}
\begin{figure*}[!htbp]
	\begin{center}
		\includegraphics[width=0.9\linewidth]{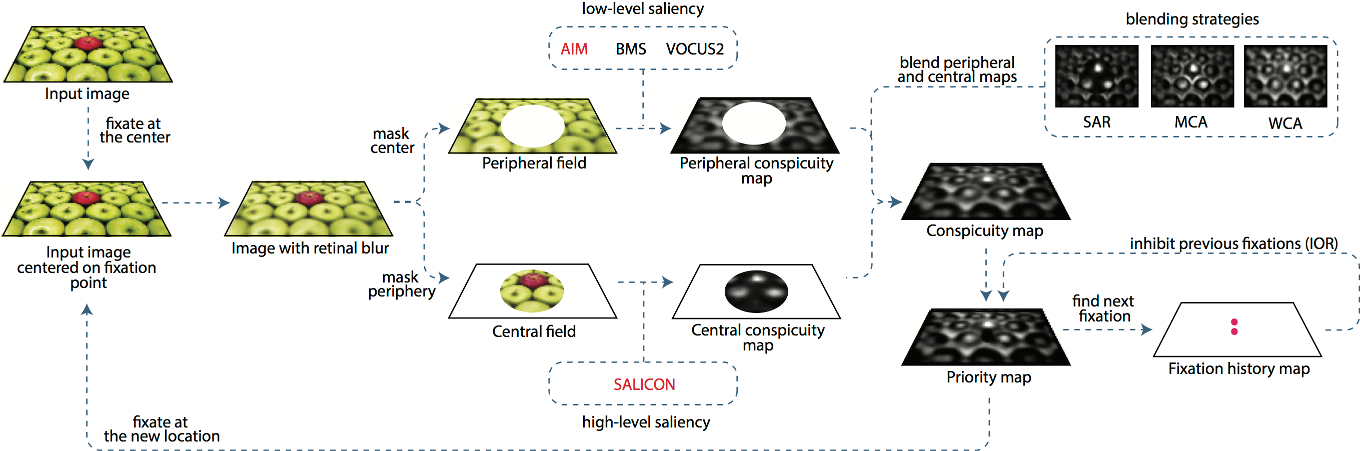} 
	\end{center}
	\vspace{-3mm}
	\caption{STAR-FC system architecture: Input images are first centrally fixated. A retinal transform is applied at the current fixation, and conspicuity is calculated within two streams: a peripheral stream which is dominated by low-level features, and a central stream which includes high-level and abstract features such as those learned by deep networks. The peripheral and central streams are then fused into a conspicuity map. The priority map combines the conspicuity map and input from a history map of all previous fixations (as well as any task-specific biases not further detailed here), providing an Inhibition of Return (IOR) mechanism. The next fixation point is selected from the maximum of the priority map, at which point the fixation is shifted to this new target location and the network repeats.
	 \label{fig:NetArchitecture}}
\end{figure*}

Our gaze control model extends and generalizes the approach initially taken by Tsotsos \etal \cite{TsotsosEtAl2016}. Their original implementation provided much of the theoretical basis for the design of our model, but was only qualitatively tested against the seminal eye tracking work of Yarbus \cite{Yarbus1967}. Without compromising the theoretical motivations of the previous work, we have modified the network structure to generalize across natural images and thereby allow quantitative testing of the model performance. Figure \ref{fig:NetArchitecture} provides a schematic of our implementation.

\begin{figure}[!htbp]
	\begin{center}
		\begin{subfigure}[b]{0.2\textwidth}
			\includegraphics[width=\textwidth]{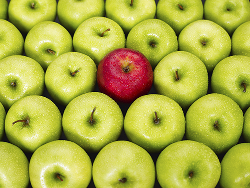}
			\caption{Original image without retinal transform}
			\label{subfig:NoTransform}
		\end{subfigure}
		\hspace{2mm}
		\begin{subfigure}[b]{0.2\textwidth}
			\includegraphics[width=\textwidth]{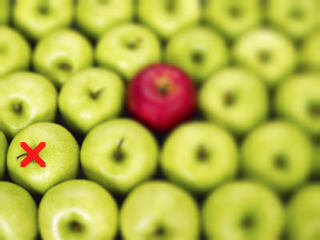}
			\caption{Fixated figure with retinal transform}
			\label{subfig:RetinalTransform}
		\end{subfigure}
	\end{center}
	\vspace{-4mm}
	\caption{An example of the retinal transform. The figure on the left shows the original input image, and the figure on the right shows the appearance of that image when fixated in the lower left location (marked with a red `X')
	 \label{fig:FoveationExample}}
	 \vspace{-4mm}
\end{figure}

The primary motivation of our architecture is to construct a set of interactive modules which introduce an iterative temporal component to fixation prediction (\ie an active approach to perception). When humans visually explore an image, each fixation is made in the context of the prior fixations, introducing a confounding difficulty for any static map attempting to predict fixation locations passively. Although it has long been pointed out that saliency maps predict fixations with differing efficacy over time \cite{TatlerGilchrist2005}, static maps predicting a probabilistic distribution of the likelihood of any particular region being fixated remain standard practice in saliency research \cite{KummererEtAl2015b, HuangEtAl2015, KummererEtAl2016}. In order to better simulate the temporal dependence of fixation order, STAR-FC processes an input image iteratively through a chain of interacting modules:

\begin{enumerate}
	\item{Retinal transform: Based on the cone distribution from \cite{GeislerPerry1998} and rod distribution from \cite{Watson2014}, we recreate the acuity field of the human eye through anisotropic blurring centered on the current fixation point. Each pixel in the image is sampled from the appropriate level of a Gaussian pyramid depending on the distance from fixation, increasing blur with distance from fixation (see Figure \ref{fig:FoveationExample} for an example). Further details are provided in the Appendix \ref{sec:RetTransform}.}
	\item{Central-peripheral split: To represent the different levels of cortical representation devoted to central versus peripheral processing, we split the image into two processing streams. Peripheral attentional capture is heavily dependent on low-level features, whereas central attentional capture is allocated at a higher level abstraction and tends to be more object-based (see \cite{TsotsosEtAl2016} for justification). In the proposed architecture this is achieved by using a bottom-up algorithm based on low-level features (e.g. AIM \cite{bruce2007saliency}, BMS \cite{ZhangSclaroff2013}, etc.) in the peripheral field and applying a CNN-based bottom-up saliency algorithm such as SALICON in the central field. The radius of the central attentional field is set to 12.5 degrees.}
	\item{Conspicuity map: The central and peripheral processing streams are recombined into a single map correspond to the original covert definition of a saliency map \cite{KochUllman1985}. Since there is no standard procedure for performing this integration, we experimented with three strategies (subsequently labeled in the text as STAR-FC\_SAR, STAR-FC\_MCA, and STAR-FC\_WCA):
	\begin{enumerate}
	\item{\emph{Separate Activation Regions} (SAR): A binary mask was applied to both the peripheral and central attentional maps to confine activations to only their respective fields. A narrow overlap region is included within which the maximum value of either the peripheral or central activation is retained (as originally proposed in \cite{TsotsosEtAl2016}).}
	\item{\emph{Maximum Central Activation} (MCA): The central attentional map is masked as in SAR, but no mask is applied to the peripheral map. Instead, the entire central region of the conspicuity map is equal to the maximum activation of either the peripheral or central maps.}
	\item{\emph{Weighted Central Activation} (WCA): The peripheral and central attentional maps are combined as follows:
	\begin{equation}
    CM_{ij}= 
\begin{cases}
    \frac{r_{c}-r_{p}}{r_{c}}\ C_{ij} + \frac{{r_p}}{r_c}P{ij},& \text{if } r_p < r_c\\
    (1 + \frac{r_p-r_c}{r_{\text{max}}-r_c})[1-g_p]P_{ij},              & \text{otherwise}
\end{cases}
	\end{equation}
	where $CM_{ij}$ is the conspicuity map value at pixel $(i,j)$, $C$ and $P$ are the central and peripheral maps, respectively, $r_c$ refers to the radius of the central field in pixels, $r_p$ is the distance to the center in pixels and $r_{\text{max}}$ is the maximum distance from the center in pixels. Here, an optional peripheral gain factor $g_p$ is introduced to increase the importance of peripheral features most affected by the retinal transform. 
	} 	
	\end{enumerate}
}	
	\item{Priority map: This combines the bottom-up activity of the conspicuity map with top-down spatial modulation. In our experiments this map only includes an inhibition of return (IOR) mechanism due to our focus on free-viewing. However, it could potentially be extended to incorporate other forms of modulation.}	
	
	\item{Fixation history map: This processing layer stores a history of previously fixated locations in image coordinates. These locations are inhibited with a circular zone of inhibition. Following \cite{TsotsosEtAl2016} the radius of IOR is set to 1.5 degrees with suppression being maximal at the point of previous fixation and linearly decreasing towards the edge. IOR decays linearly within 100 fixations. In this paper IOR is applied by subtracting the fixation history map from the priority map.}
	\item{Saccade control: This module is responsible for finding a new target within the priority map using a WTA scheme, shifting the gaze to a new location (by re-applying the retinal transform centered on the new fixation coordinates), as well as updating the fixation history map.}
\end{enumerate}

As mentioned, our work has been heavily influenced by the proposed control architecture in \cite{TsotsosEtAl2016}, but makes a number of important modifications and extensions. The original approach utilizes manually-derived face filters in the central field, specific to the single test image used for illustration. In order to generalize performance across natural images, we remove the custom face filters and instead incorporate, as part of the central field, a deep convolutional neural network (CNN), namely the SALICON saliency detection model \cite{HuangEtAl2015}. In our implementation we use a C++ conversion of the OpenSALICON \cite{Thomas2016}. 

Our choice of using a CNN-based saliency algorithm is motivated by the idea that such saliency models can be viewed as processing incoming visual information analogous to a full forward pass through the visual hierarchy in order to produce high-level feature abstraction and object-based conspicuity allocation \cite{KummererEtAl2017}. This is consistent with the theoretical aims of the central field put forth in \cite{TsotsosEtAl2016}. SALICON was specifically chosen due to the availability of an open-source implementation, but our formulation is agnostic to the specific saliency representations used in its construction.

Furthermore, we experiment with several bottom-up saliency algorithms to demostrate the effect of using different low-level features for computing peripheral attentional maps. In addition to AIM, which was also used in \cite{TsotsosEtAl2016}, we tested BMS \cite{ZhangSclaroff2013, ZhangSclaroff2016} and VOCUS2 \cite{FrintropEtAl2015}. 

Despite the fact that BMS significantly outperforms AIM on the CAT2000 dataset using the traditional saliency metrics of the MIT Saliency Benchmark \cite{BylinskiEtAl}, when utilized in the peripheral component of STAR-FC both BMS and VOCUS2 achieve much worse fidelity to human fixation patterns than is achieved with AIM, leading us to focus most of our tests on optimizing the AIM-based architecture. 

Finally, we define two additional strategies for combining the central and peripheral attentional maps aiming to alleviate the sharp border between the central and peripheral fields. This allows our architecture to more smoothly transition its activity across the visual field. See Appendices \ref{sec:SaccadeAmplitudes} and \ref{sec:TrajScores} for further information on the different STAR-FC variants we tested.

Although virtually any saliency algorithm can be used within the proposed architecture, both the choice of saliency algorithms for the central/peripheral fields and strategy for combining them have a dramatic effect on the produced fixation sequences. This will be discussed in more detail in Section \ref{sec:Results}.  
 
\subsection{Fixation Dataset}
\label{sub:Dataset}
We evaluated model performance over the CAT2000 dataset\footnote{A number of fixations included in the individual sequences of observers for the CAT2000 dataset end up going outside the bounds of the image. In order to prevent spurious comparisons with out of bound fixations while still ensuring cohesive sequences, we groomed the CAT2000 data by truncating any sequence which went out of bounds to the final in-bounds fixation location. If this truncation left the sequence with fewer than ten total fixations, it was discarded completely. Of 36000 total recorded fixation sequences, this criterion led to the elimination of 6257 sequences.} \cite{BorjiItti2015}. This dataset was chosen due to several positive attributes: it contains twenty different image categories (thereby representing a wide spectrum of visual stimuli), as well as one of the widest fields of view which we are aware of for a free-viewing eye tracking dataset (approximately $45^\circ$). Larger fields of view better approximate natural scene exploration, and are also likely to be more greatly impacted by considerations of retinal anisotropy and motoric bias than a comparable dataset gathered over a narrow field of view.

\subsection{Evaluation Metrics}
\label{sub:Metrics}
One major challenge in this work was determining the best method for evaluation. The output of our fixation control model is not directly comparable to that of saliency algorithms designed to predict human fixations, as we output a sparse set of explicitly predicted locations rather than a smooth map which can be treated as a probability distribution for likely fixation points over an image \cite{KummererEtAl2015b}. However, as mentioned in Section \ref{sub:Applications}, there are applications for which an explicit sequence of fixation points is preferable to a probabilistic heat-map which lacks temporal structure.

Given that the innovation of our work rests on providing an explicit, temporally ordered fixation sequence rather than on a novel representation of saliency, we focus on evaluation metrics which reflect the spatiotemporal structure of sequences. In order to compare against the static maps which are the standard output of saliency algorithms, we sampled fixation sequences from the maps by applying an iterative WTA procedure. IOR was applied to each selected location using the same parameters as those of our fixation control model. This technique is consistent with previous work which samples loci of attention from saliency maps \cite{IttiKoch2000}.

Although saccade amplitude distributions provide a relatively coarse measure with which to compare fixation sequences (as there is no representation of positional differences over the visual field), they do provide a representation of the motoric bias in the prediction. An early criticism of saliency algorithms was that they fail to account for inherent motor biases in how humans move their eyes \cite{TatlerVincent2009}, and it has been suggested that this motor bias could implicitly contribute to the persistent challenge of center bias in saliency research \cite{WlokaTsotsos2016}. We therefore examine this aspect of model function in Section \ref{sub:Amplitudes}, demonstrating a much more human-like distribution of saccade amplitude with our model than is found from the predictions of sampled from static saliency maps.

To more explicitly explore the prediction performance of our model, we utilize trajectory-based scoring methods. These metrics focus on measuring the deviation between two spatiotemporal sequences. Trajectory comparison is a common problem in a wide range of fields, and can often rely on a number of different constraints or assumptions. Three common classifications of trajectory metrics are \emph{network-constrained}, \emph{shape-based}, and \emph{warping-based} \cite{BesseEtAl2016}. Network-constrained methods rely on an underlying path structure (such as a road network), and were therefore not appropriate for our purposes. However, both shape-based (which measure the spatial structure of trajectories) and warping-based (which take into account the temporal structure as well as the spatial) can provide meaningful insight for saccadic sequences, and we therefore utilized the following set in order to provide a comprehensive sense of performance (trajectory-based score results are found in Section \ref{sub:Scores}): 

\begin{itemize}[noitemsep]
	\item{\emph{Euclidean Distance (ED)}: ED is one of the most common and basic warping-based trajectory metrics, and is calculated by matching two sequences in temporal order and computing the average pairwise distance between corresponding fixation points.}
	\item{\emph{Fr{\'e}chet Distance (FD)}: FD is sometimes referred to as the `dog-walking distance'; it represents the maximum distance at any given point in time over the length of two trajectories.}
	\item{\emph{Hausdorff Distance (HD)}: HD is the maximum distance of a point in one sequence to the nearest point in a second sequence. Unlike ED and FD, HD is purely spatial and does not take sequence order into account.}
\end{itemize}

\begin{figure}[!htbp]
	\begin{center}
		\begin{subfigure}[b]{0.35\textwidth}
			\includegraphics[width=\textwidth]{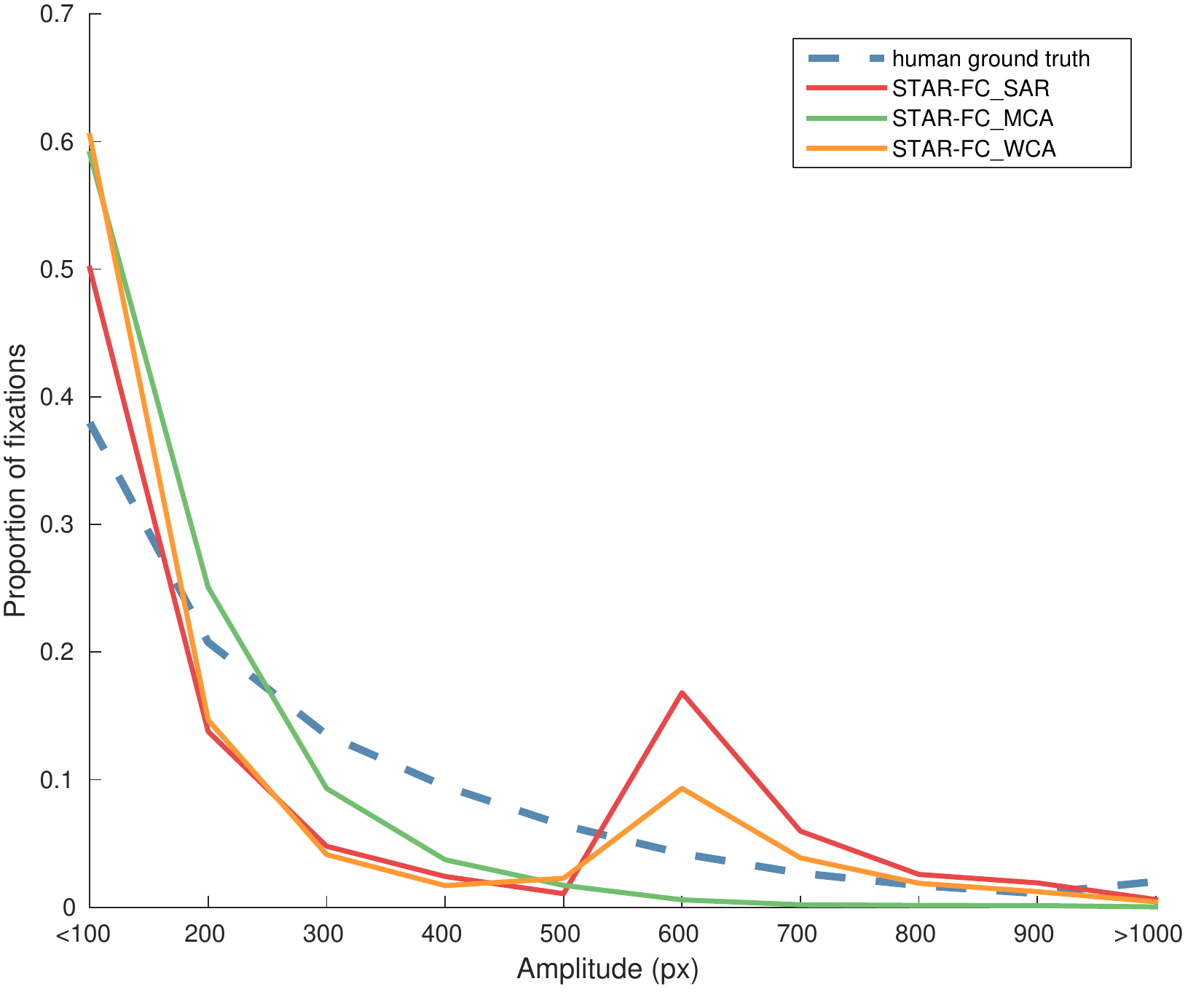}
			\caption{STAR-FC variants compared to humans \label{subfig:FC_amps}}
		\end{subfigure}
		\begin{subfigure}[b]{0.35\textwidth}
			\includegraphics[width=\textwidth]{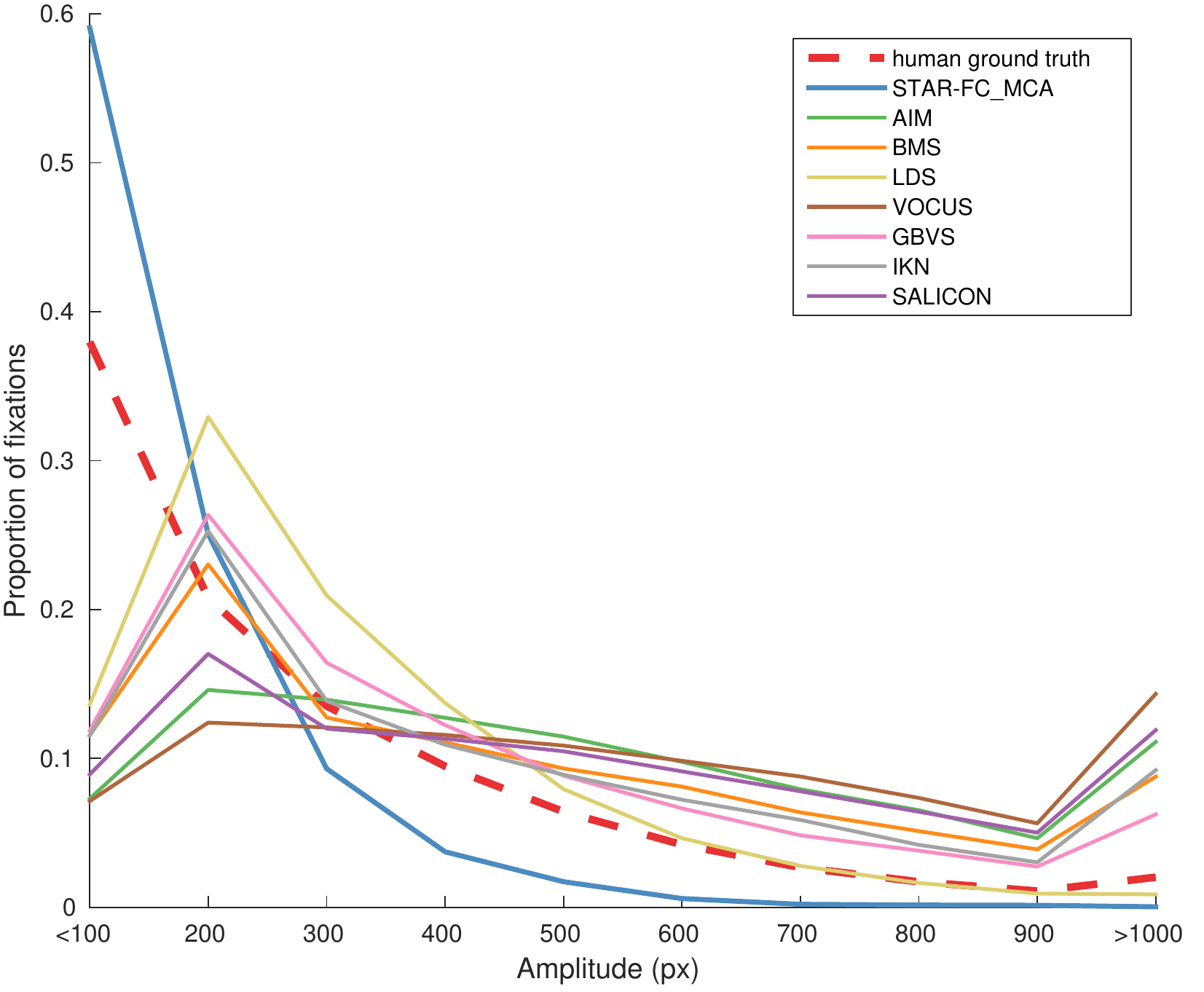}
			\caption{Traditional saliency algorithms, STAR-FC, and human distributions \label{subfig:FCvsSal_amps}}
		\end{subfigure}
	\end{center}
	\vspace{-4mm}
	\caption{A graphical depiction of the saccadic amplitude distributions over the CAT2000 dataset. Saccade lengths were assigned to bins of pixel ranges and the proportion of saccades falling in each bin are shown in the figures. Figure (a.) shows the effect of the different STAR-FC configurations on the resultant saccadic amplitude distribution (contrasted with the human distribution shown with a dashed line) Figure (b.) shows the distributions of a selection of traditional saliency algorithms contrasted with the MCA variant of STAR-FC and the human distribution. 
	 \label{fig:Amplitudes}}
	\vspace{-4mm}
\end{figure}

\section{Results}
\label{sec:Results}

We compare the performance of our STAR-FC with a range of established saliency models: AIM \cite{BruceTsotsos2007}, BMS \cite{ZhangSclaroff2013}, GBVS \cite{HarelKochPerona2007}, LDS \cite{ShuEtAl2017}, SALICON \cite{HuangEtAl2015,Thomas2016}, SSR \cite{SeoMilanfar2009}, and VOCUS2 \cite{FrintropEtAl2015}. For additional comparisons see Appendices \ref{sec:SaccadeAmplitudes} through \ref{sec:TrajScores}.

\subsection{Spatial Distributions}
\label{sub:Amplitudes}

\begin{figure*}[!tbp]
	\begin{center}
		\begin{subfigure}[b]{0.125\textwidth}
			\includegraphics[width=\textwidth]{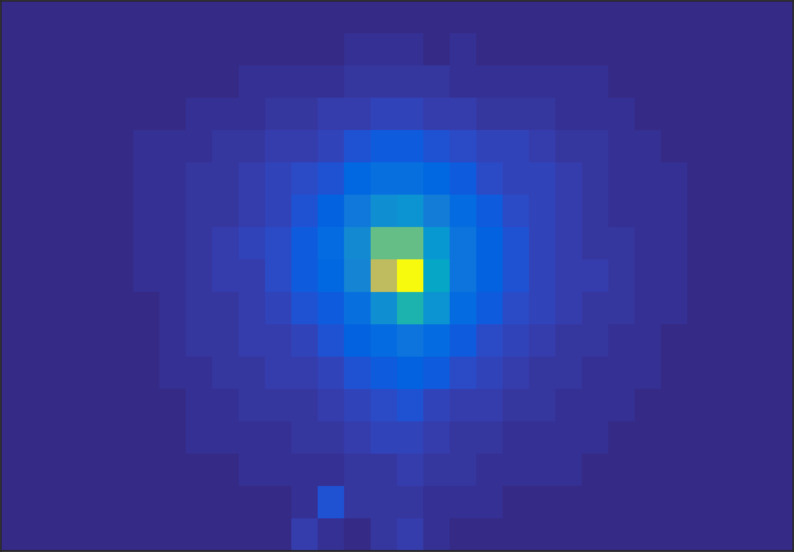}
			\caption{Human \\ (MSE Score)}
		\end{subfigure}
		\begin{subfigure}[b]{0.125\textwidth}
			\includegraphics[width=\textwidth]{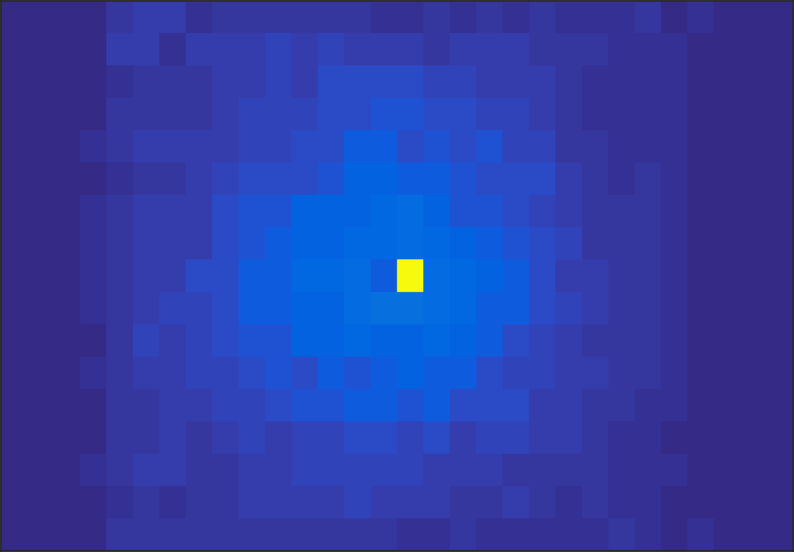}
			\caption{STAR-FC, \\ (0.002)}
		\end{subfigure}
		\begin{subfigure}[b]{0.125\textwidth}
			\includegraphics[width=\textwidth]{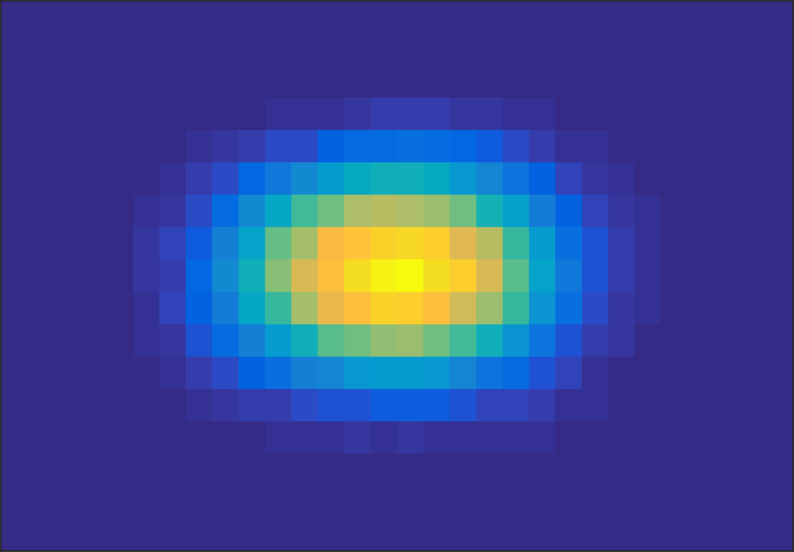}
			\caption{LDS, \\ (0.027)}
		\end{subfigure}
		\begin{subfigure}[b]{0.125\textwidth}
			\includegraphics[width=\textwidth]{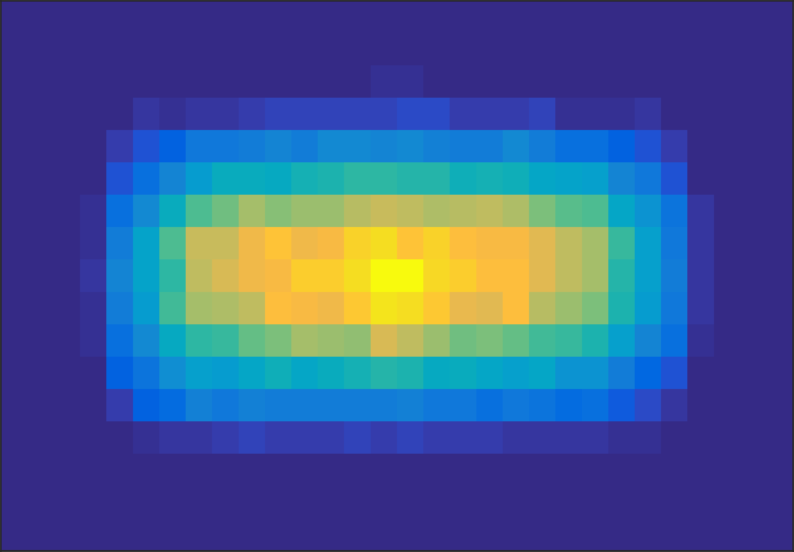}
			\caption{GBVS, \\ (0.072)}
		\end{subfigure}
		\begin{subfigure}[b]{0.125\textwidth}
			\includegraphics[width=\textwidth]{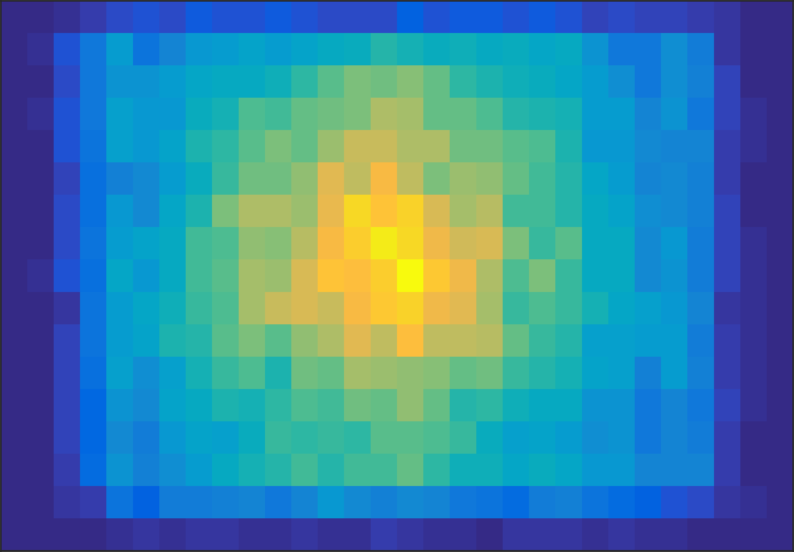}
			\caption{SALICON, \\ (0.122)}
		\end{subfigure}
	\end{center}
	\vspace{-6mm}
	\caption{2D histograms of fixation locations over the CAT2000 dataset. Mean-squared-error (MSE) scores between model and human distributions are shown in parentheses under each model name; as can be seen, STAR-FC is an order of magnitude closer to the human distribution than the closest competing saliency model.
		\label{fig:2D_Hists}}
	\vspace{-4mm}
\end{figure*}

Saccadic amplitude distributions are shown in Figure \ref{fig:Amplitudes}. As can be seen in Figure \ref{subfig:FC_amps}, the original central-peripheral integration strategy of Separate Activation Regions (SAR) used in \cite{TsotsosEtAl2016} has a tendency to create a bimodal distribution not seen in the human data. This is likely due to the fact that the retinal anisotropy creates a biased gradient to the output of both the central and peripheral fields, meaning that near the border of the two the central field is weakest and the peripheral field is strongest. In order to facilitate a smoother transition of activation across the visual field, we tested two other integration strategies (described in Section \ref{sub:Architecture}): Maximum Central Activation (MCA) and Weighted Central Activation (WCA).

Our motivation to allow for the low-level feature representation of the peripheral map to affect the central region but not the other way around is based on the fact that there do appear to be fundamental perceptual limitations in object perception and feature binding within peripheral vision \cite{StrasburgerEtAl2011}, whereas low-level features do seem to have a persistent role in attentional guidance \cite{KummererEtAl2017}.

Despite blending peripheral and central activations in a smoothly merging fashion, the WCA strategy leads to an activation pattern remarkably similar to the original SAR strategy. This is likely due to the fact that a weighted blending will usually lead to penalizing the chances of both algorithms within the mid-central region to attract attention unless they both happen to achieve a high score, essentially requiring a target to attract both high and low level attention simultaneously.

The closest distribution pattern to that of humans was achieved by the MCA integration strategy, and it is therefore the variant reported in Figure \ref{fig:FC_vs_Saliency} and Table \ref{tab:Results}. Although it does match the human distribution more closely than WCA and SAR variants, MCA appears to over-emphasize short saccades, having a much shallower tail than seen in the distribution of human observers. As previously mentioned, one likely contribution to this over-emphasis is the difficulty of many algorithms which have not been explicitly designed or trained to deal with signal degradation to function effectively across the retinal transform.

In contrast to the STAR-FC amplitude distributions, virtually all static saliency maps are skewed in the opposite direction with distributions which are much flatter than those seen with human data. Many algorithms do retain a small preference for shorter saccades, but this could also be an outcome of compositional bias in the underlying images. 2D histograms of fixation location produced with $64 \times 64$ sized blocks across the full CAT2000 dataset are shown for humans along with the MCA variant of STAR-FC as well as several representative saliency algorithms in Figure \ref{fig:2D_Hists}. As can be seen, there does appear to be a consistent spatial bias toward the center of the image which, at least in part, likely represents the underlying composition of the dataset images. Likewise, the saliency algorithms with the closest spatial distribution to the human distribution do tend to have a greater propensity for shorter saccades (as seen in Figure \ref{fig:Amplitudes}).

\subsection{Trajectory Scores}
\label{sub:Scores}

\begin{figure}[!htbp]
	\begin{center}
		\begin{subfigure}[b]{0.35\textwidth}
			\includegraphics[width=\textwidth]{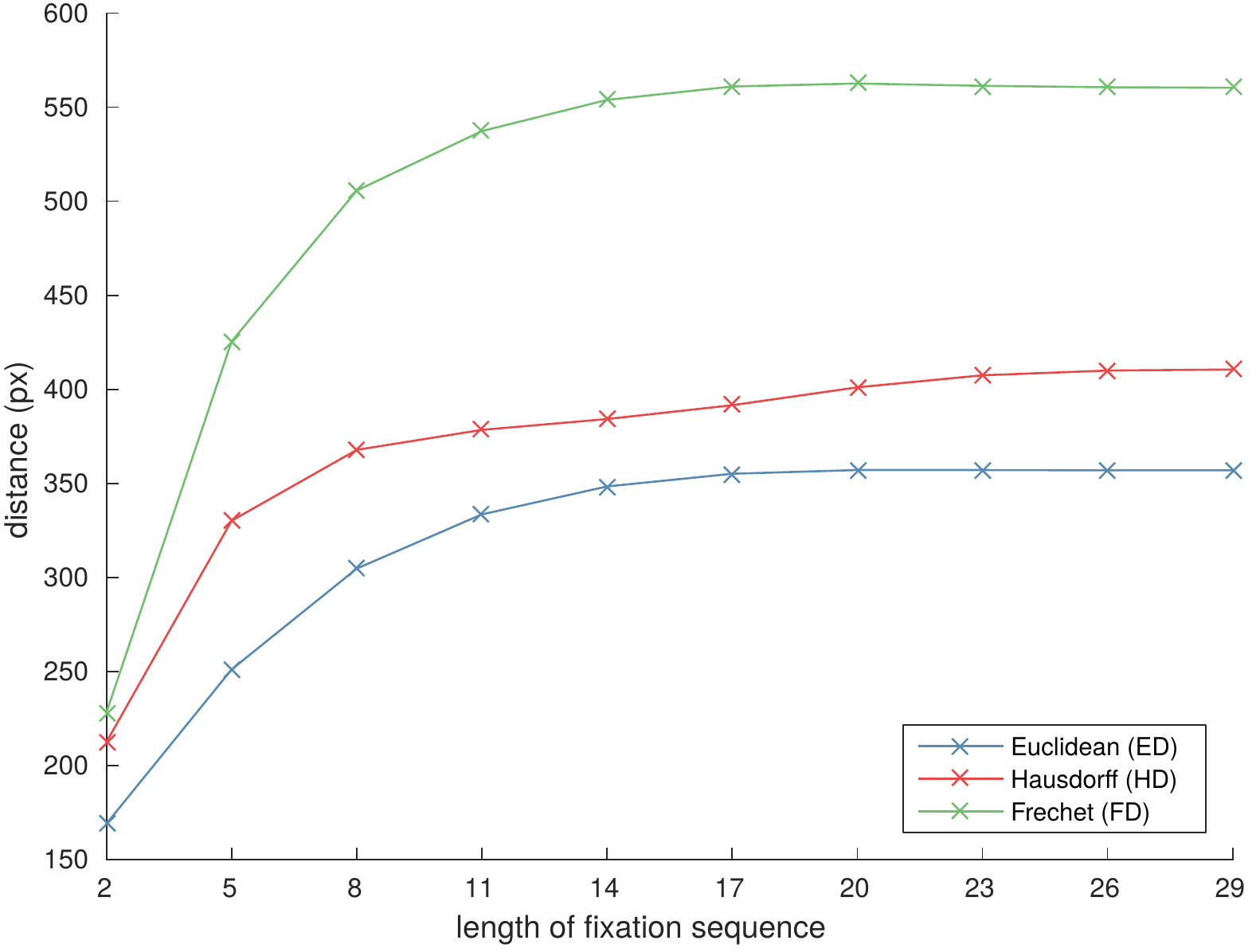}
			\caption{Average scores with sequence length \label{subfig:Human_Sequence_Length}}
		\end{subfigure}
		\begin{subfigure}[b]{0.35\textwidth}
			\includegraphics[width=\textwidth]{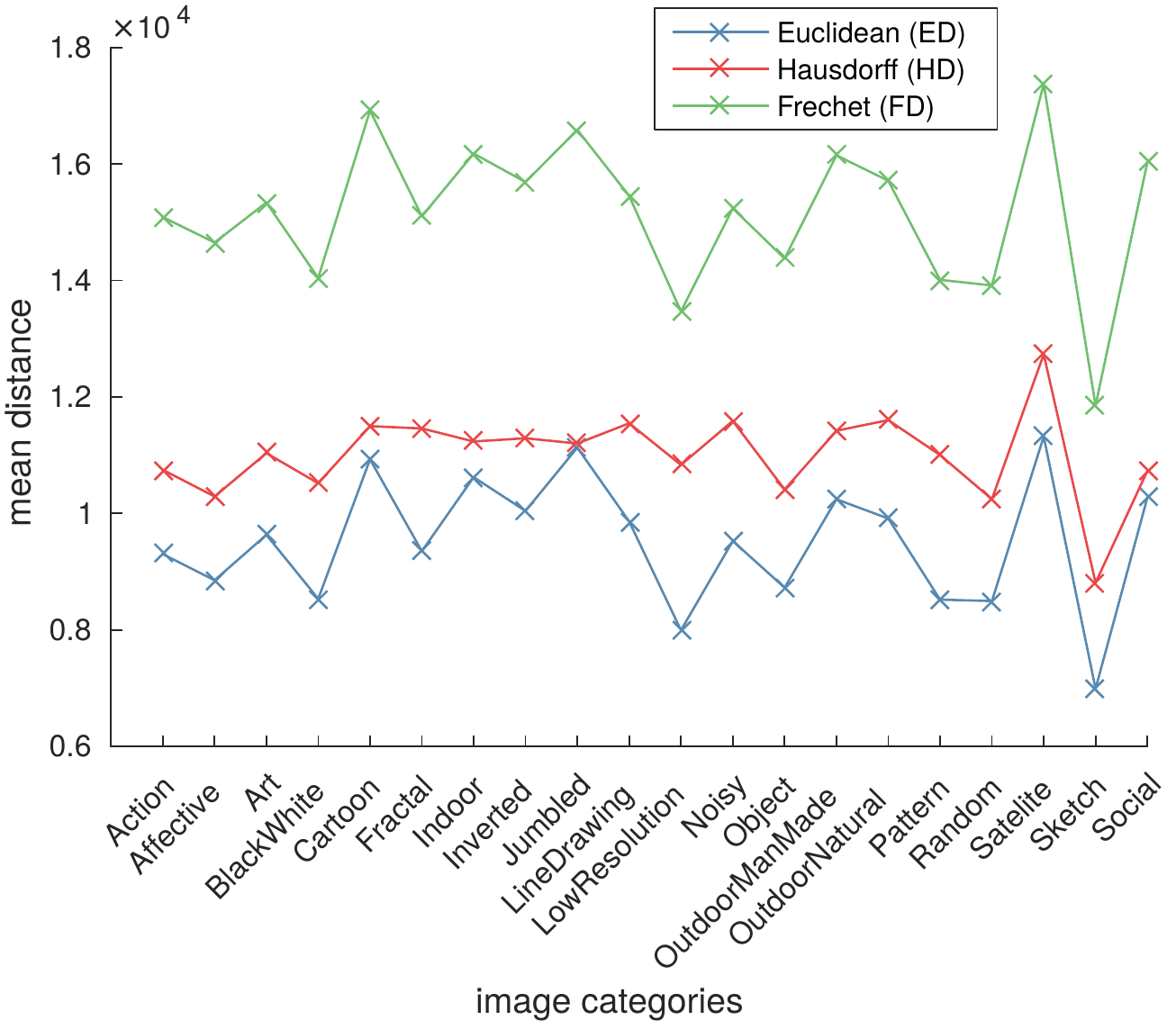}
			\caption{Average total sequence score by category \label{subfig:Human_Category}}
		\end{subfigure}
	\end{center}
	\vspace{-3mm}
	\caption{Average scores computed for all metrics over pair-wise matches of human sequences. As can be seen in (a.), as sequence length increases observer agreement tends to diverge, leading to a saturation in score values for each metric. Figure (b.) shows average sequence score per category, showing agreement with \cite{BorjiItti2015} about which categories tend to have greatest inter-observer consistency.
	 \label{fig:Human_scores}}
	\vspace{-3mm}
\end{figure}

\begin{figure*}[!htbp]
	\begin{center}
		\begin{subfigure}[b]{0.3\textwidth}
			\includegraphics[width=\textwidth]{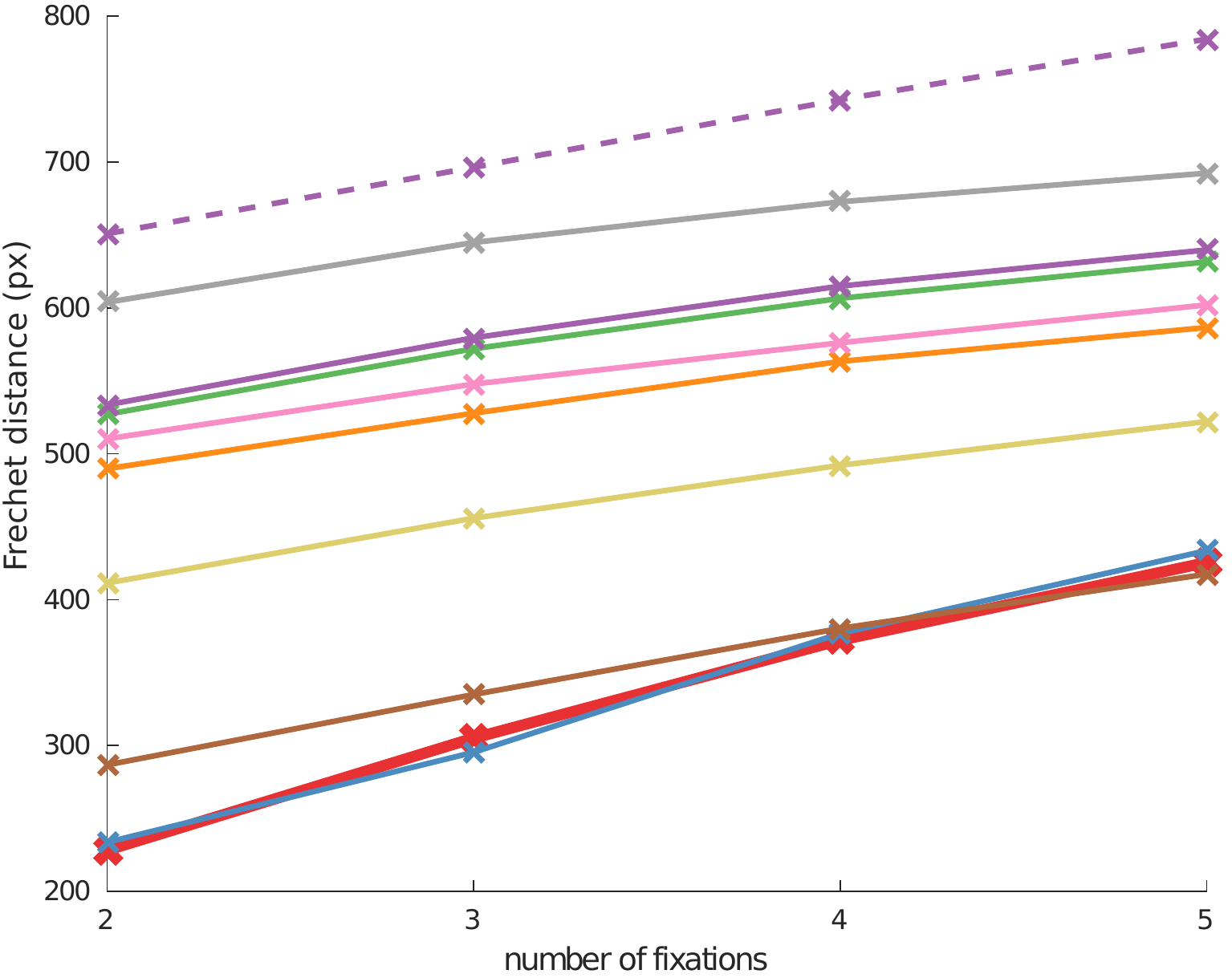}
			\caption{Mean FD}
		\end{subfigure}
		\begin{subfigure}[b]{0.3\textwidth}
			\includegraphics[width=\textwidth]{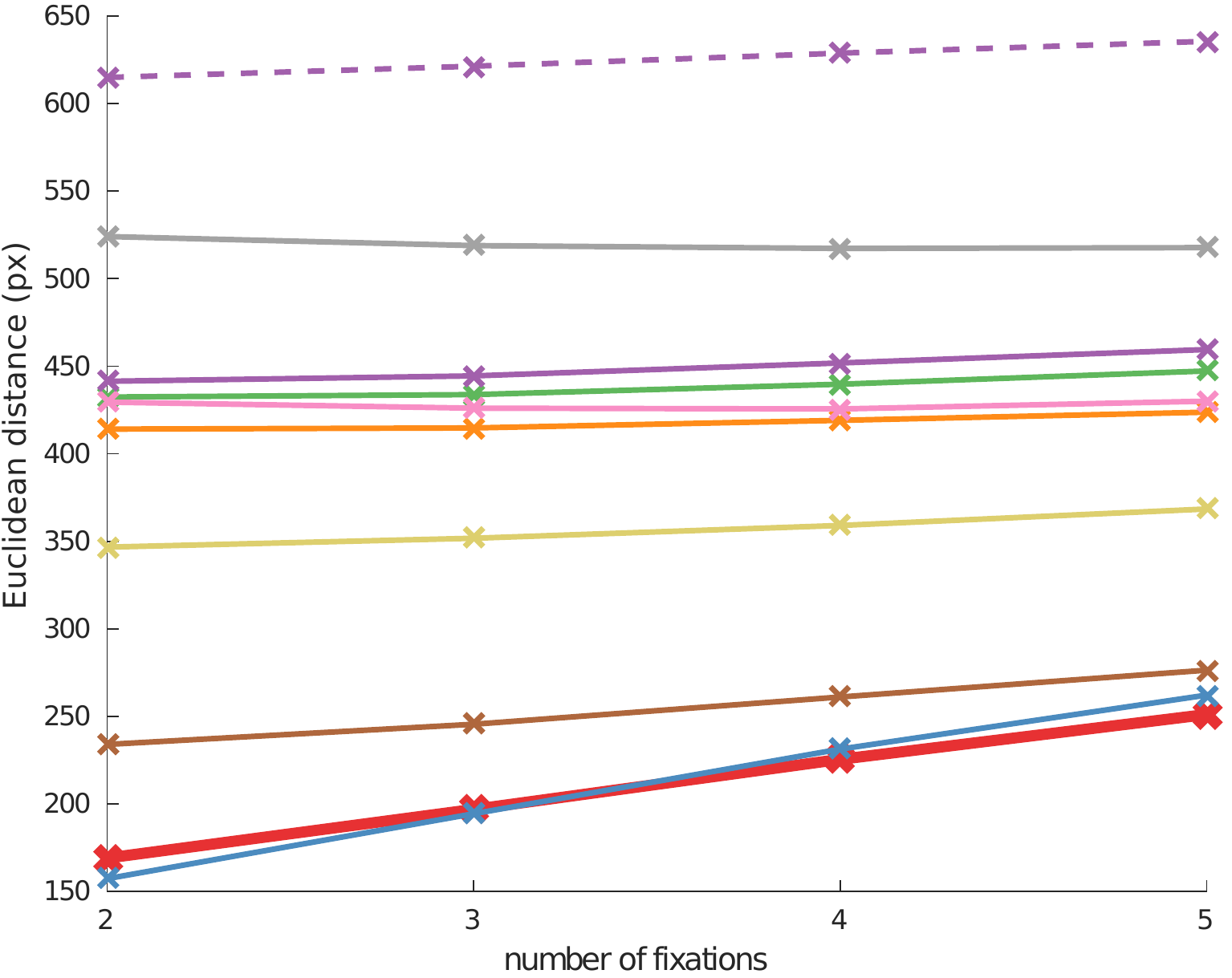}
			\caption{Mean ED}
		\end{subfigure}
		\begin{subfigure}[b]{0.3\textwidth}
			\includegraphics[width=\textwidth]{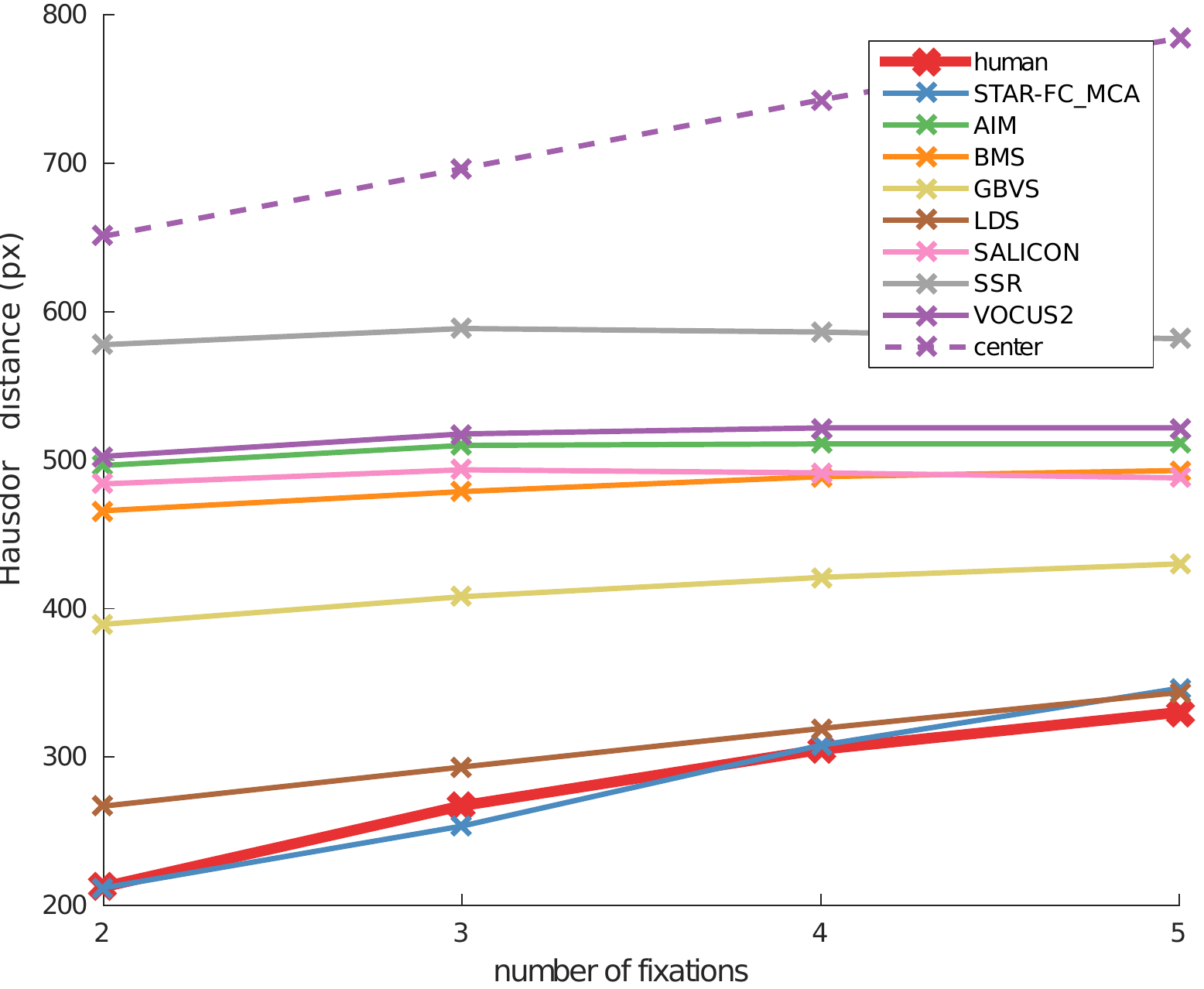}
			\caption{Mean HD}
		\end{subfigure}
	\end{center}
	\vspace{-3mm}
	\caption{A comparison of fixation prediction scores for static saliency maps and STAR-FC. A sequence formed by always picking the center pixel is shown in a dashed line to provide a performance baseline.
	 \label{fig:FC_vs_Saliency}}
\end{figure*}

\begin{table}[!bt]
    \begin{center}
    \begin{tabular}{| c | c | c | c | c |}
        \hline
        Model & AUC ED & AUC HD & AUC FD & MSE\\ \hline \hline
        Human & 632 & 844 & 1004 & 0 \\ \hline 
        STAR-FC & \textbf{630} & \textbf{841} & \textbf{1006} & \textbf{0.002} \\ \hline
        LDS & 762 & 918 & 1067 & 0.027 \\ \hline
        GBVS & 1068 & 1239 & 1415 & 0.072 \\ \hline
        BMS & 1253 & 1447 & 1629 & 0.102 \\ \hline
        SALICON & 1281 & 1471 & 1680 & 0.122 \\ \hline
        AIM & 1313 & 1525 & 1758 & 0.161 \\ \hline
        VOCUS2 & 1347 & 1551 & 1781 & 0.183 \\ \hline
        SSR & 1557 & 1755 & 1966 & 0.183 \\ \hline
        center & 1875 & 2156 & 2156 & 0.008 \\
        \hline
    \end{tabular}
    \caption{Algorithm performance. Area-under-the-curve (AUC) scores are reported over the first five fixations for each plot in Figure \ref{fig:FC_vs_Saliency}. The last column shows the mean-square-error for the spatial histogram of predicted fixations versus the distribution of human fixations over the entire dataset. Note that our model (in bold) matches the inter-subject error of human observers. \label{tab:Results}}
    \vspace{-7mm}
    \end{center}
\end{table}

Figure \ref{fig:Human_scores} shows the results of computing pair-wise scores across all combinations of human sequences for each image from the CAT2000 dataset. Figure \ref{subfig:Human_Sequence_Length} shows that the different trajectory metrics all tend to drift toward a saturated value; ED, FD, and HD all get larger as sequences diverge through time. Additionally, it has been shown that saliency tends to correlate best with early fixations \cite{TatlerGilchrist2005}, and both saliency correlation and inter-observer consistency degrade largely after the first five fixations. We therefore restrict our analysis to only this interval. Analysis of the full sequences may be found in Appendix \ref{sec:TrajScores}.

Figure \ref{subfig:Human_Category} shows the category-wise average total sequence scores per category. Here we can see that the trajectory metrics largely agree with the analysis done by Borji and Itti \cite{BorjiItti2015} on which categories have the greatest inter-observer consistency (such as Sketch, Low Resolution, and Black and White), and which categories tend to have poor inter-observer consistency (such as Satellite, Jumbled, and Cartoon).

We compare STAR-FC against a wide selection of saliency algorithms in Figure \ref{fig:FC_vs_Saliency}, showing that STAR-FC consistently achieves trajectory scores more in line with human sequences over the critical range of the earliest fixations, followed by LDS and GBVS (see Table \ref{tab:Results} for numerical scores). In fact, STAR-FC is the only model which is able to achieve near-parity with the natural heterogeneity found within human observers across all three trajectory metrics employed. LDS, the best performing saliency algorithm, has $20.6\%$ increased ED, $8.8\%$ increased HD, and $6.3\%$ increased FD, with all other saliency algorithms increasing in average distance from the human trajectories.

As is made clear in Figure \ref{fig:2D_Hists}, human fixations over CAT2000 are strongly biased toward the center, a distribution which is well-matched by STAR-FC. The best performing saliency algorithms (LDS \cite{ShuEtAl2017} and GBVS \cite{HarelKochPerona2007}) likewise have correspondingly stronger biases toward predicting fixations near the image center. We therefore also tested the "center" model, which is simply a sequence which always selects the central pixel for every fixation. This selection will minimize the upper error bound for all trajectory metrics, and can be qualitatively thought of as a similar performance baseline to a centered Gaussian for more traditional saliency metrics \cite{JuddEtAl2009}. Nevertheless, as Figure \ref{fig:FC_vs_Saliency} shows, the center model consistently achieves the worst score in all metrics, confirming that while a centrally focused distribution of fixation locations is appropriate for the CAT2000 dataset, it is not a sufficient characteristic to score well.

\section{Conclusion}
\label{sec:Conclusion}

Our Fixation Control model provides a powerful tool for predicting explicit fixation sequences.  demonstrating fidelity to human fixation patterns equivalent to that of using one person's fixation sequence to predict another. This performance is significantly better than what can be achieved by sequence sampling from static saliency maps (see Table \ref{tab:Results}), as the STAR-FC model is the only fixation prediction method capable of achieving average trajectory scores on par with human inter-observer comparisons. STAR-FC will therefore allow improved performance in saliency applications relying on explicit fixation prediction, including for commercial \cite{VAS_Sample2015} and science communication \cite{HaroldEtAl2016} purposes. In addition to its performance, our model is also constructed to provide a descriptive model of fixation control, allowing further research into the interaction of the different cognitive control architectures which link gaze to higher order visual cognition \cite{TsotsosKruijne2014}.

While it is clear that retinal anisotropy has a significant effect on human visual performance, very few computational algorithms are developed with the aim of dealing with anisotropic acuity. This creates a significant challenge to accurately detect and ascribe conspicuity values across the visual field, and our model's incorporation of retinal anisotropy represents an interesting platform for exploring this area of research.

Additionally, free-viewing over static images represents only a very narrow range of task for which fixation prediction provides valuable information. Fixation prediction over video and under task demands are highly challenging domains for which explicit fixation control may prove extremely valuable.

\FloatBarrier
{\small
\bibliographystyle{ieee}
\bibliography{FixControlBib}
}

\begin{appendices}
\section{Retinal Transform}
\label{sec:RetTransform}
Our implementation follows the same steps as outlined in \cite{TsotsosEtAl2016} with few minor changes. In particular, we refitted the generalized Gamma distribution to better adjust with respect to viewing parameters and reimplemented the interpolation function in CUDA.

As with most foveation algorithms, our approach starts by building a Gaussian pyramid and then for each pixel the appropriate level of the pyramid is sampled depending on how far the pixel is from the current gaze point. 
The level of the pyramid to sample from is computed as follows
\begin{equation}
L_{x,y} = \frac{\frac{\pi}{180}(\text{atan}((D_{\text{rad}}+\text{dotpitch})\frac{1}{D_{\text{view}}}) - \text{atan}((D_{\text{rad}}-\text{dotpitch})\frac{1}{D_{\text{view}}}))}{(\epsilon_{2} (\alpha*(EC+\epsilon_{2}))).*log(\frac{1}{CT_{0}})}
\end{equation}
where $D_{\text{rad}}$ is the radial distance between the point $(x,y)$ and the current gaze point, $EC$ is the eccentricity from the fovea center for point $(x,y)$ in degrees, dotpitch is the size of the pixel of the monitor in meters and $D_{\text{view}}$ refers to the viewing distance. $\alpha$, $\epsilon_2$ and $CT_{0}$ are constants from \cite{GeislerPerry1998}. In this equation the numerator represents the maximum spatial frequency that can be represented at the given distance from the current gaze point and the denominator is the maximum spatial resolution that can be resolved by the eye. 

However, \cite{GeislerPerry1998} concentrated on cone vision. For a more complete and biologically consistent results we also provide the option of augmenting the cone model with rod vision following \cite{TsotsosEtAl2016} using the generalized Gamma distribution as proposed in \cite{Watson2014}. Since in \cite{Watson2014} cell counts were used to fit a distribution function, we adjust the parameters to convert it to the units that we use, namely the levels of a Gaussian pyramid. Therefore, we set the parameters of the generalized Gamma distribution as follows: $\alpha=2.46, \beta=121.8, \gamma=0.77, \sigma=861.27$ and $\mu=-1$. To find corresponding levels of the pyramid for each pixel we compute the generalized gamma distribution for $EC$ and plug it into the equation (1) as the denominator. Due to the severe drop off of cones in the far periphery, the rods function is of greater impact the larger the field of view, but has relatively little impact over the eccentricities present in the CAT2000 dataset.

Finally we compute the foveated image using a bi-cubic interpolation routine for 3D volumes to sample the required level of the pyramid for each pixel. Our code is a CUDA reimplementation of the \texttt{ba\_interp3} function \footnote{\url{https://www.mathworks.com/matlabcentral/fileexchange/21702-3d-volume-interpolation-with-ba-interp3--fast-interp3-replacement}}.

Note that we compute cone distribution for each color channel separately, but rod distribution only for the intensity channel (the image is first converted to YCrCb color space), since rods are achromatic. When including rods, we use a default proportion of contribution from the rods and cones functions of $30\%$ and $70\%$, respectively, to the final transformed image. 

The viewing conditions in all our experiments match the experimental conditions reported for the CAT2000 dataset \cite{BorjiItti2015}, namely all stimuli span 45 degrees and the viewing distance is set to 1.06 m.

\section{Saccade amplitudes}
\label{sec:SaccadeAmplitudes}
\begin{figure*}[ht]
	\begin{center}
		\begin{subfigure}[b]{0.3\textwidth}
			\includegraphics[width=\textwidth]{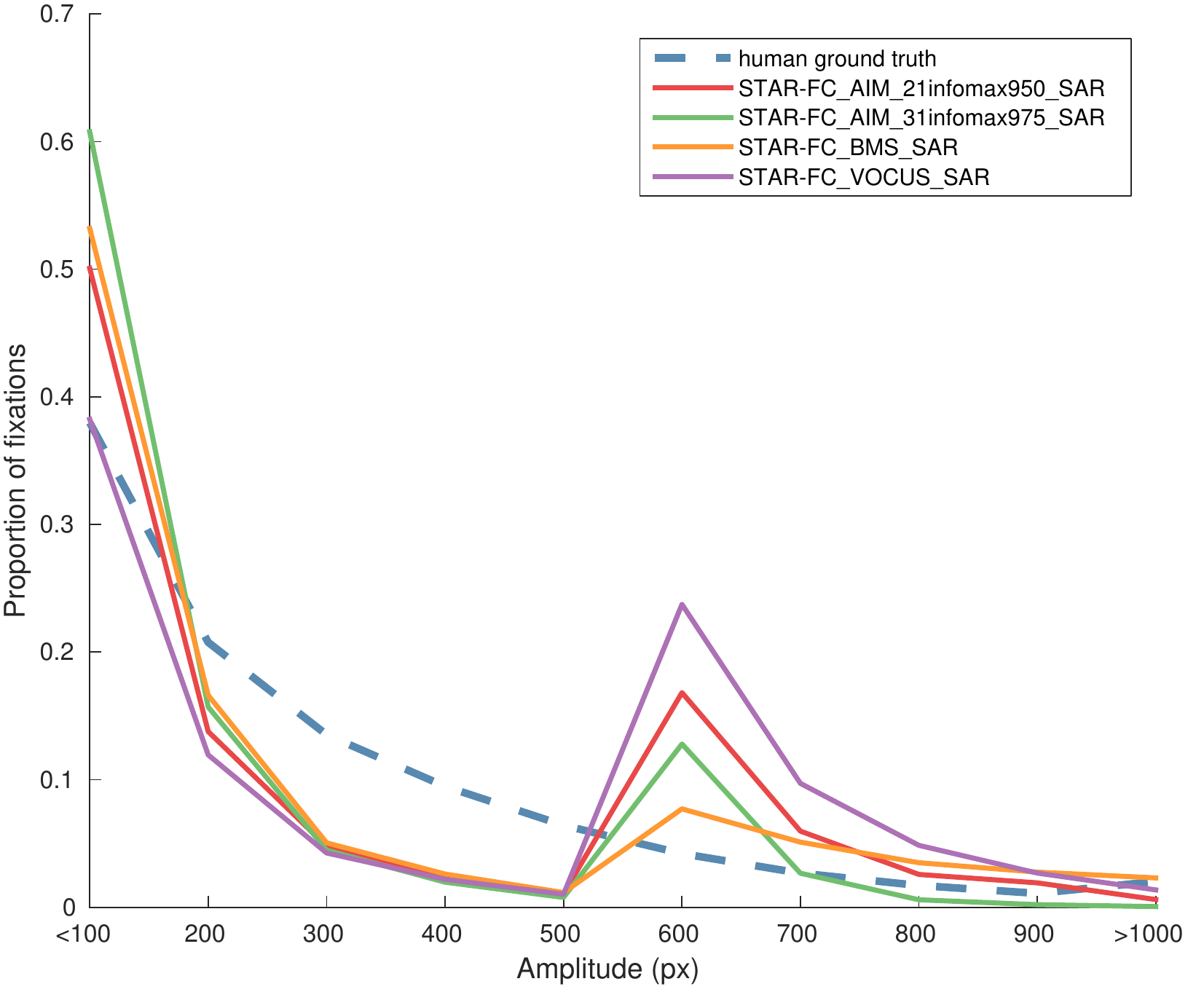}
			\caption{SAR\label{fig:Saccade_amplitudes_SAR}}
		\end{subfigure}
		\begin{subfigure}[b]{0.3\textwidth}
			\includegraphics[width=\textwidth]{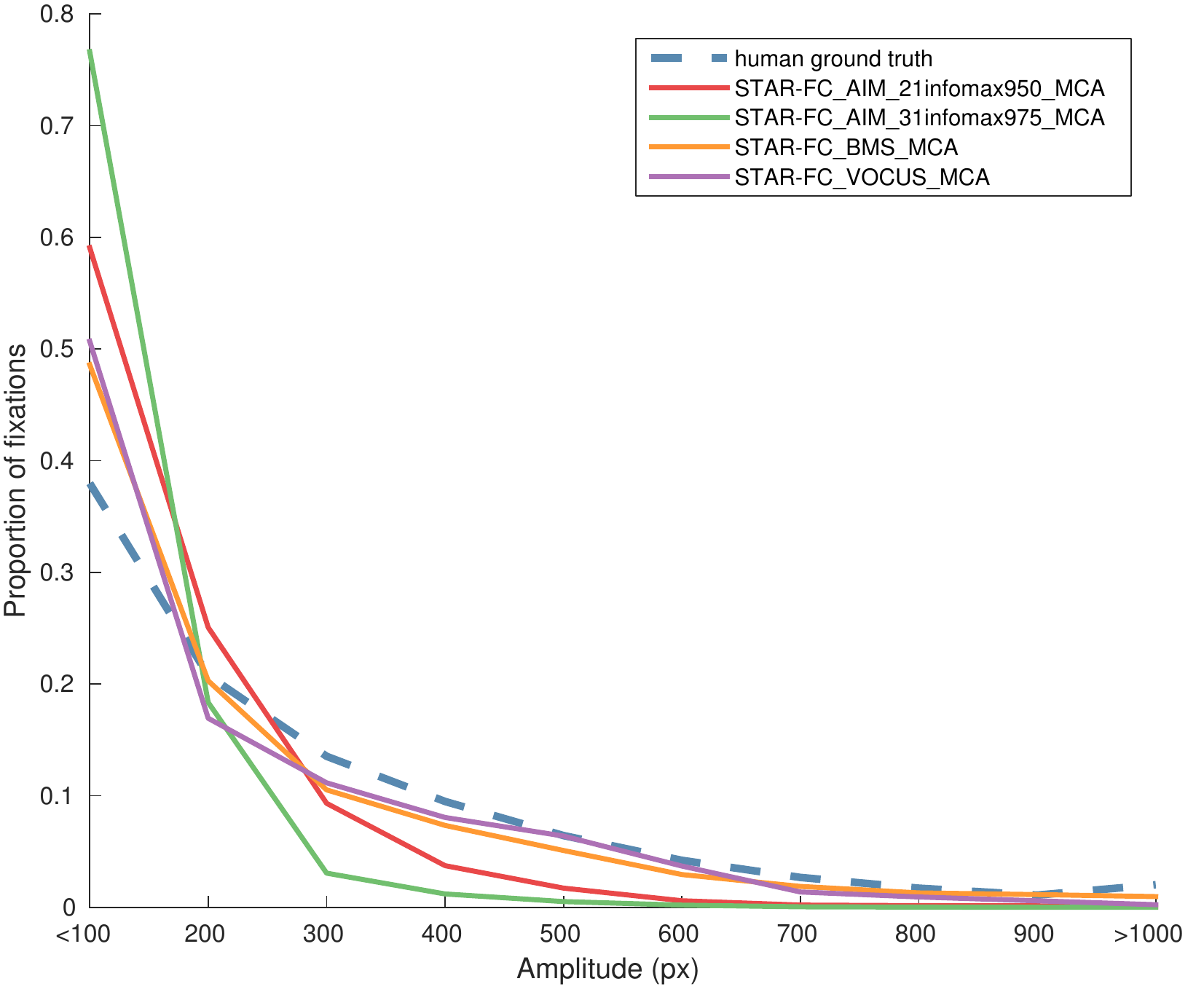}
			\caption{MCA\label{fig:Saccade_amplitudes_MCA}}
		\end{subfigure}
		\begin{subfigure}[b]{0.3\textwidth}
			\includegraphics[width=\textwidth]{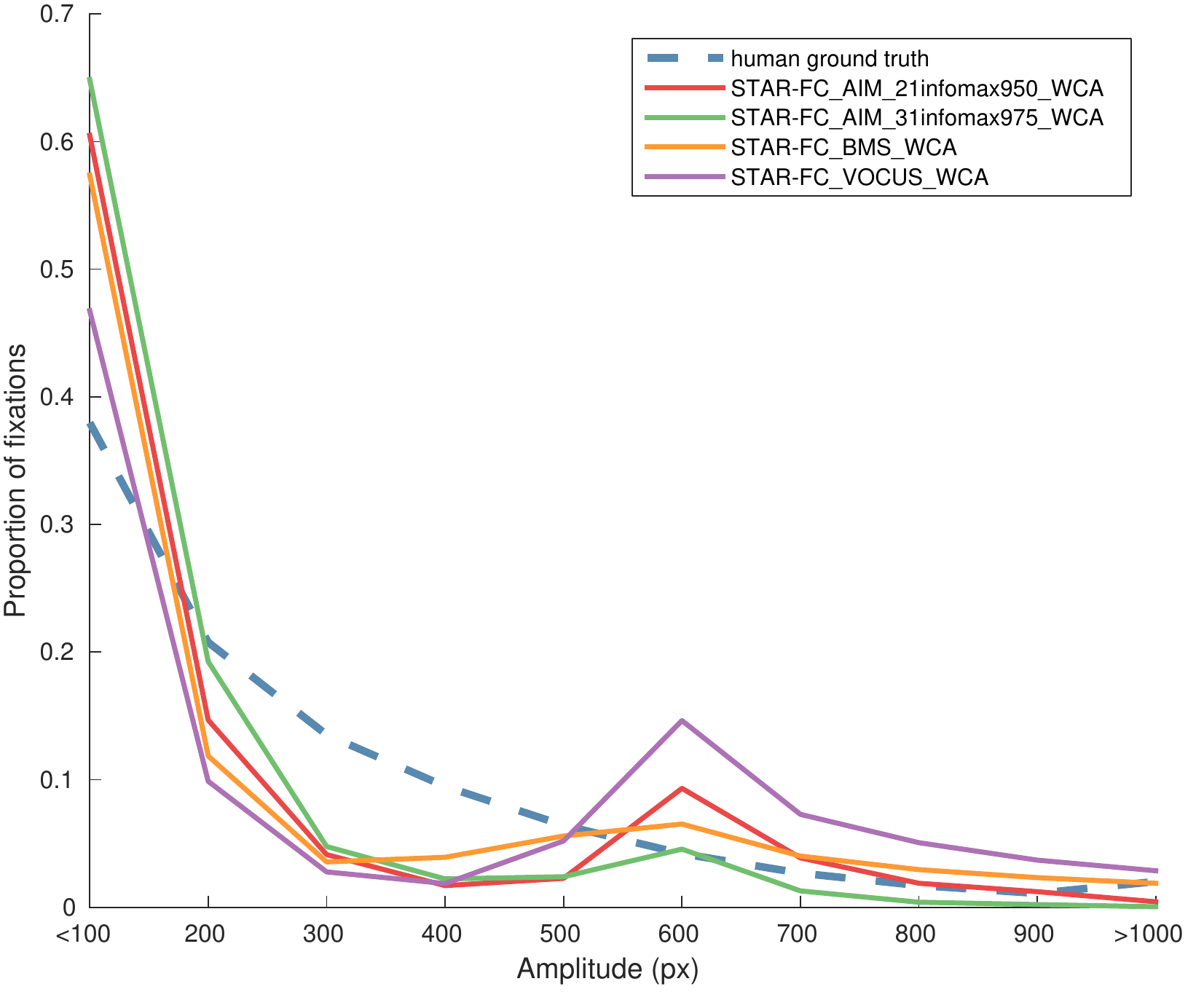}
			\caption{WCA\label{fig:Saccade_amplitudes_WCA}}
		\end{subfigure}
	\end{center}
	\vspace{-4mm}
	\caption{Plots of fixation amplitudes demonstrating the effects of different strategies for combining peripheral and central fields of STAR-FC (SAR, MCA and WCA), and different bottom-up saliency algorithms in the peripheral field (AIM, VOCUS and BMS).
		\label{fig:Saccade_amplitudes}}
	\vspace{-4mm}
\end{figure*}

\begin{figure}[ht]
	\begin{center}
		\includegraphics[width=0.9\linewidth]{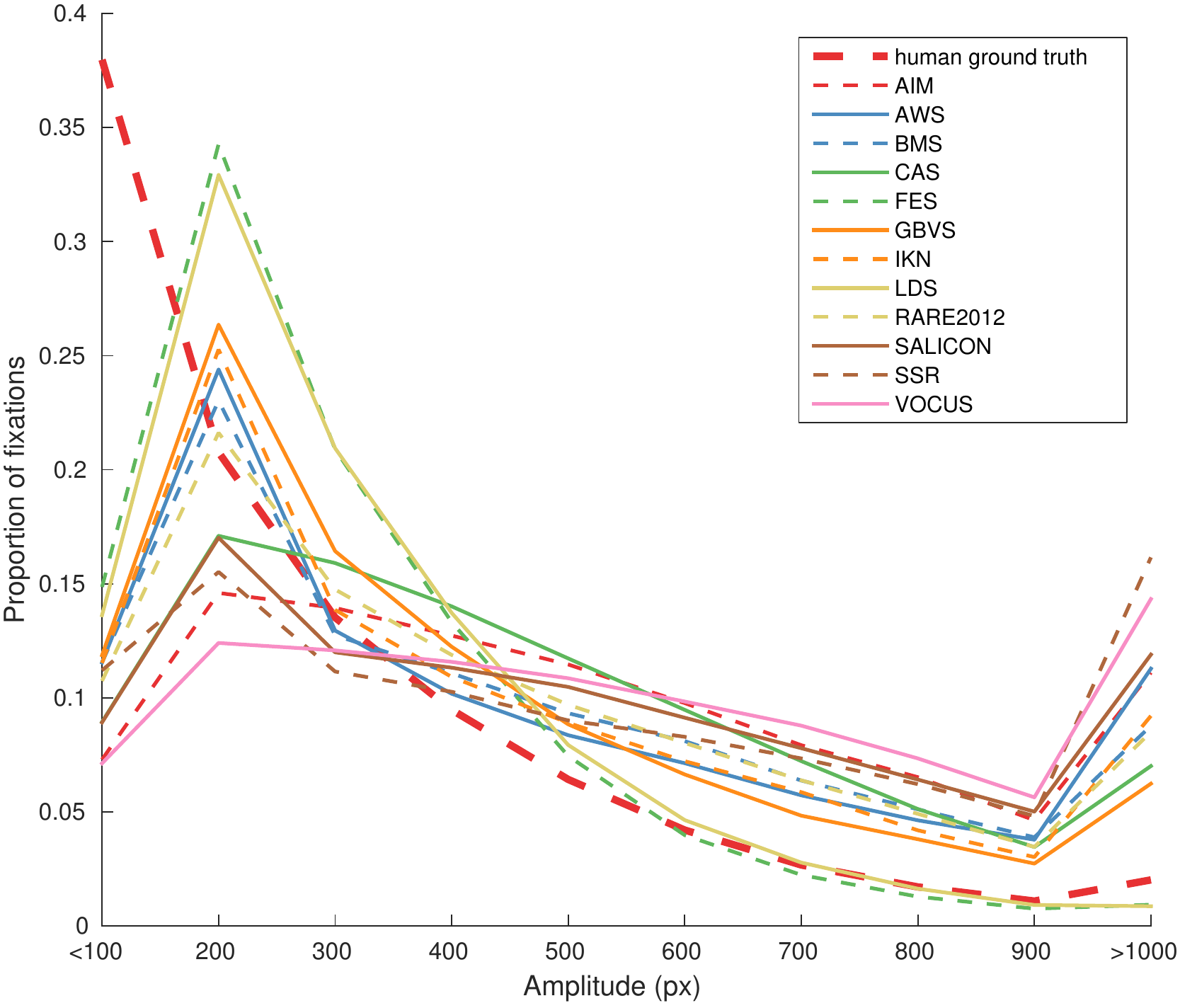} 
	\end{center}
	\vspace{-3mm}
	\caption{Fixation amplitudes for 12 state-of-the-art bottom-up saliency algorithms.
	 \label{fig:SaliencySaccadeAmplitudes}}
	\vspace{-4mm}
\end{figure}

\figref{fig:Saccade_amplitudes} shows plots of the fixation amplitudes that demonstrate the effect of using different saliency algorithms in the periphery and blending strategies. 
Both SAR and WCA blending strategies (\figref{fig:Saccade_amplitudes_SAR} and \figref{fig:Saccade_amplitudes_WCA}) lead to a pronounced spike in the distribution, which approximately corresponds to the diameter of the central field. MCA, on the other hand, produces a more even distribution of the amplitudes (\figref{fig:Saccade_amplitudes_MCA}).
\figref{fig:SaliencySaccadeAmplitudes} shows fixation amplitudes for all tested bottom-up saliency algorithms. Note that all of them greatly underestimate the number of short saccades ($<$ 100 px) and generally have a much flatter fall off than the human ground truth distribution.

\section{2D Histograms of Fixations}
\label{sec:2DFixHist}
\begin{figure*}[!htbp]
	\begin{center}
		\begin{subfigure}[b]{0.17\textwidth}
			\includegraphics[width=\textwidth]{images/2D_fix_histograms/human_fixations_2D_hist.png}
			\caption{Human \\ (MSE Score)}
		\end{subfigure}
		\begin{subfigure}[b]{0.17\textwidth}
			\includegraphics[width=\textwidth]{images/2D_fix_histograms/cat2k_AIM_21infomax950_MCA_fixations_2D_hist.png}
			\caption{STAR-FC, \\ (0.002)}
		\end{subfigure}
		\begin{subfigure}[b]{0.17\textwidth}
			\includegraphics[width=\textwidth]{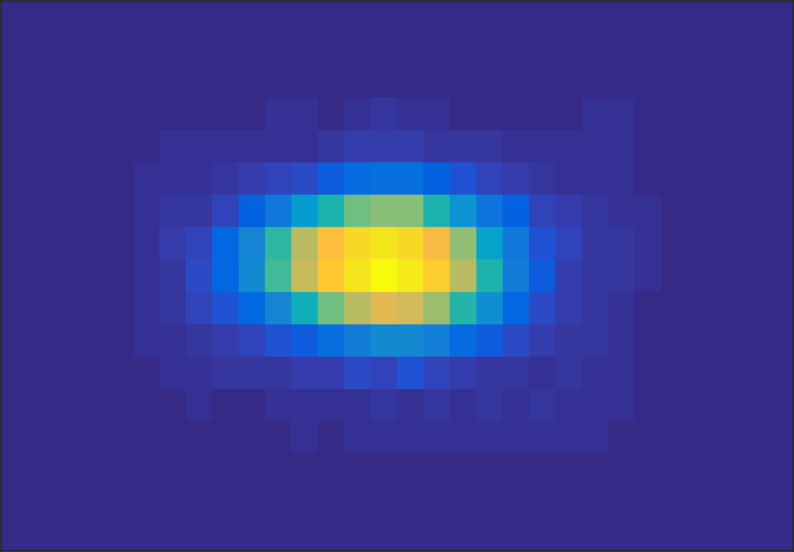}
			\caption{FES, \\ (0.01)}
		\end{subfigure}
		\begin{subfigure}[b]{0.17\textwidth}
			\includegraphics[width=\textwidth]{images/2D_fix_histograms/LDS_fixations_2D_hist.png}
			\caption{LDS, \\ (0.026)}
		\end{subfigure}
		\begin{subfigure}[b]{0.17\textwidth}
			\includegraphics[width=\textwidth]{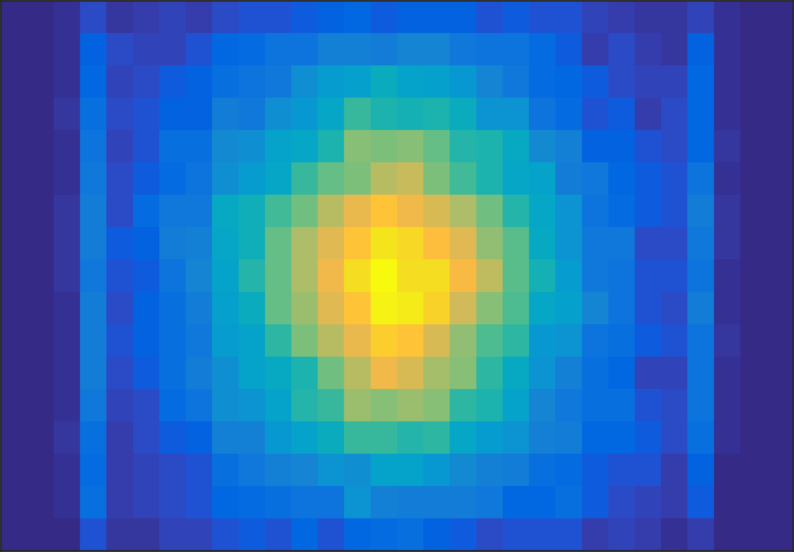}
			\caption{CAS, \\ (0.067)}
		\end{subfigure}
				\begin{subfigure}[b]{0.17\textwidth}
			\includegraphics[width=\textwidth]{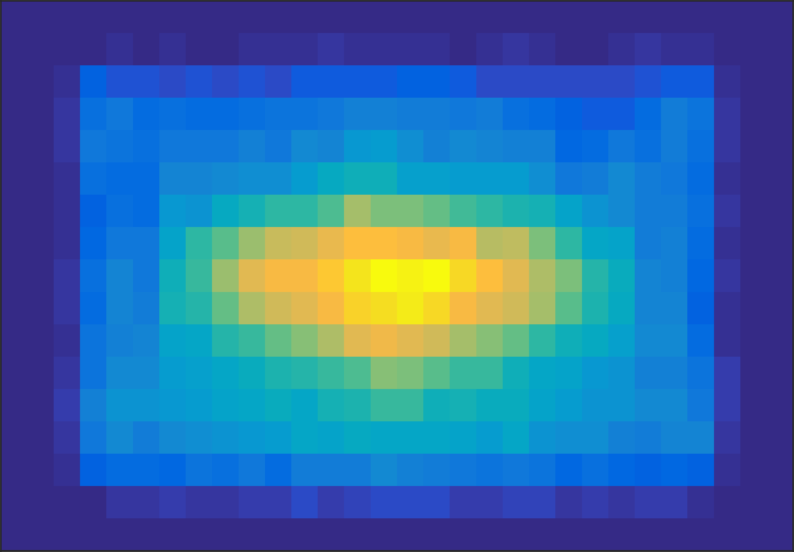}
			\caption{IKN \\ (0.07)}
		\end{subfigure}
		\begin{subfigure}[b]{0.17\textwidth}
			\includegraphics[width=\textwidth]{images/2D_fix_histograms/GBVS_fixations_2D_hist.png}
			\caption{GBVS, \\ (0.07)}
		\end{subfigure}
		\begin{subfigure}[b]{0.17\textwidth}
			\includegraphics[width=\textwidth]{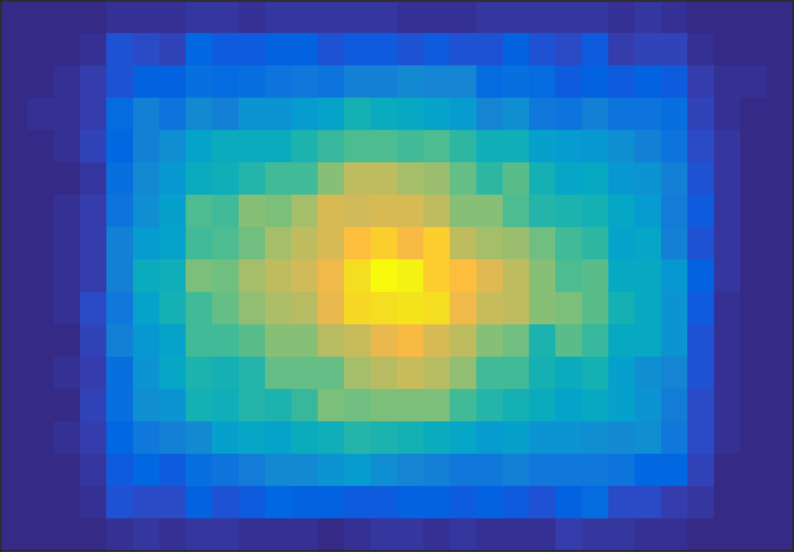}
			\caption{RARE2012, \\ (0.09)}
		\end{subfigure}
		\begin{subfigure}[b]{0.17\textwidth}
			\includegraphics[width=\textwidth]{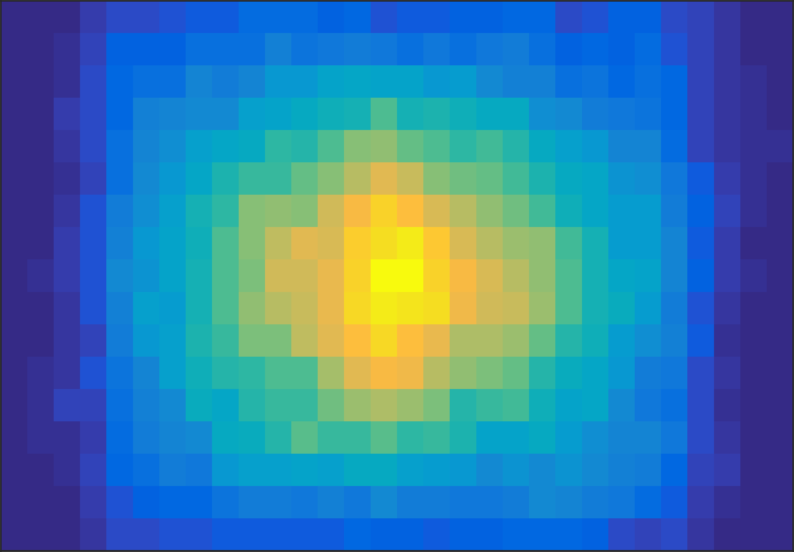}
			\caption{BMS, \\ (0.102)}
		\end{subfigure}
		\begin{subfigure}[b]{0.17\textwidth}
			\includegraphics[width=\textwidth]{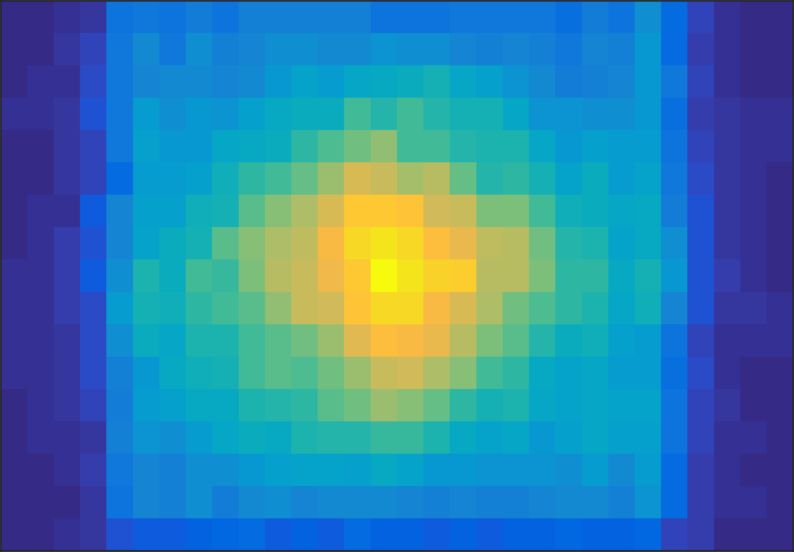}
			\caption{AWS, \\ (0.104)}
		\end{subfigure}
				\begin{subfigure}[b]{0.17\textwidth}
			\includegraphics[width=\textwidth]{images/2D_fix_histograms/SALICON_fixations_2D_hist.png}
			\caption{SALICON, \\ (0.12)}
		\end{subfigure}
		\begin{subfigure}[b]{0.17\textwidth}
			\includegraphics[width=\textwidth]{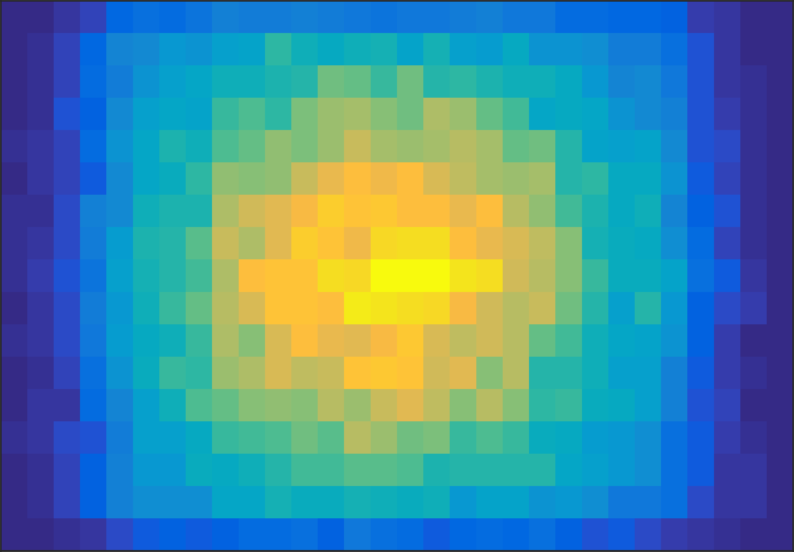}
			\caption{AIM, \\ (0.16)}
		\end{subfigure}
		\begin{subfigure}[b]{0.17\textwidth}
			\includegraphics[width=\textwidth]{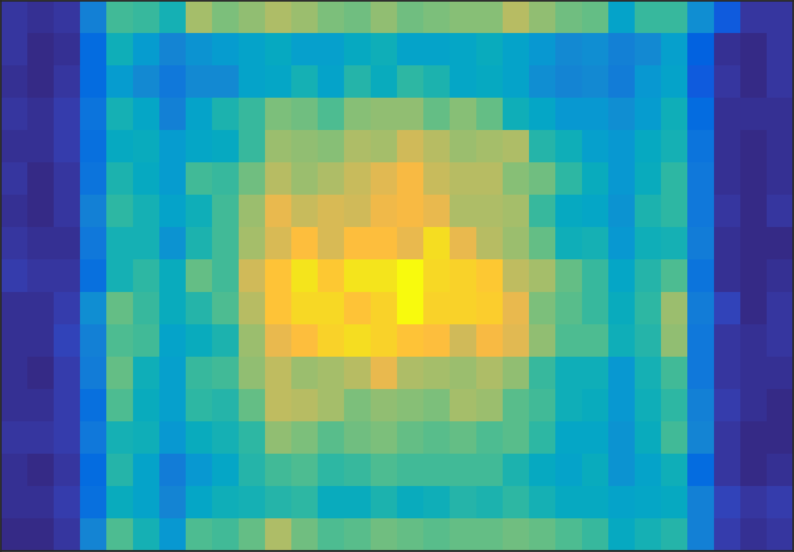}
			\caption{SSR, \\ (0.183)}
		\end{subfigure}
		\begin{subfigure}[b]{0.17\textwidth}
			\includegraphics[width=\textwidth]{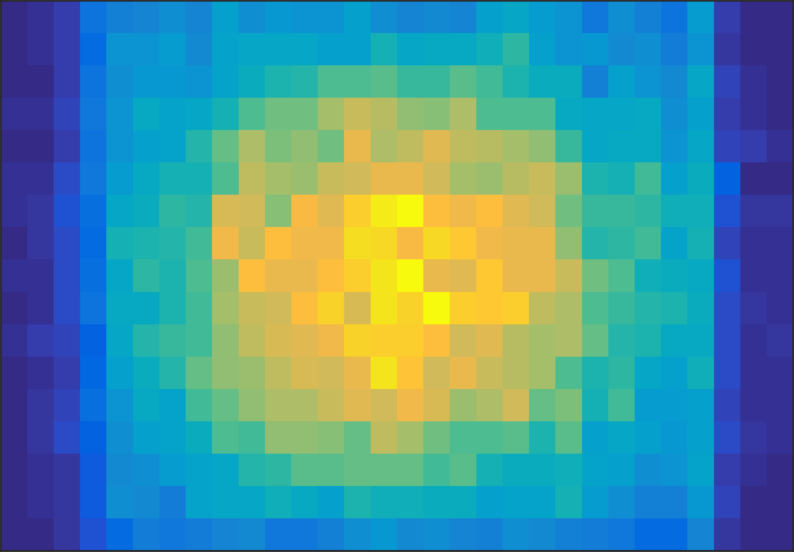}
			\caption{VOCUS, \\ (0.183)}
		\end{subfigure}
	\end{center}
	\vspace{-6mm}
	\caption{2D histograms of fixation locations over the CAT2000 dataset for all tested bottom-up saliency algorithms. Mean-squared-error (MSE) scores between model and human distributions are shown in parentheses under each model name. The algorithms are sorted by MSE in ascending order, starting with STAR-FC (AIM with 21infomax950 basis and MCA blending strategy), which is an order of magnitude closer to the human distribution than the best bottom-up algorithm (FES).
		\label{fig:2D_Hists_apdx}}
\end{figure*}

\figref{fig:2D_Hists_apdx} shows 2D histograms of fixations for all saliency algorithms with MSE scores.

\section{Trajectory Scores}
\label{sec:TrajScores}
\begin{figure*}[!htbp]
	\begin{center}
		\begin{subfigure}[b]{0.75\textwidth}
			\includegraphics[width=\textwidth]{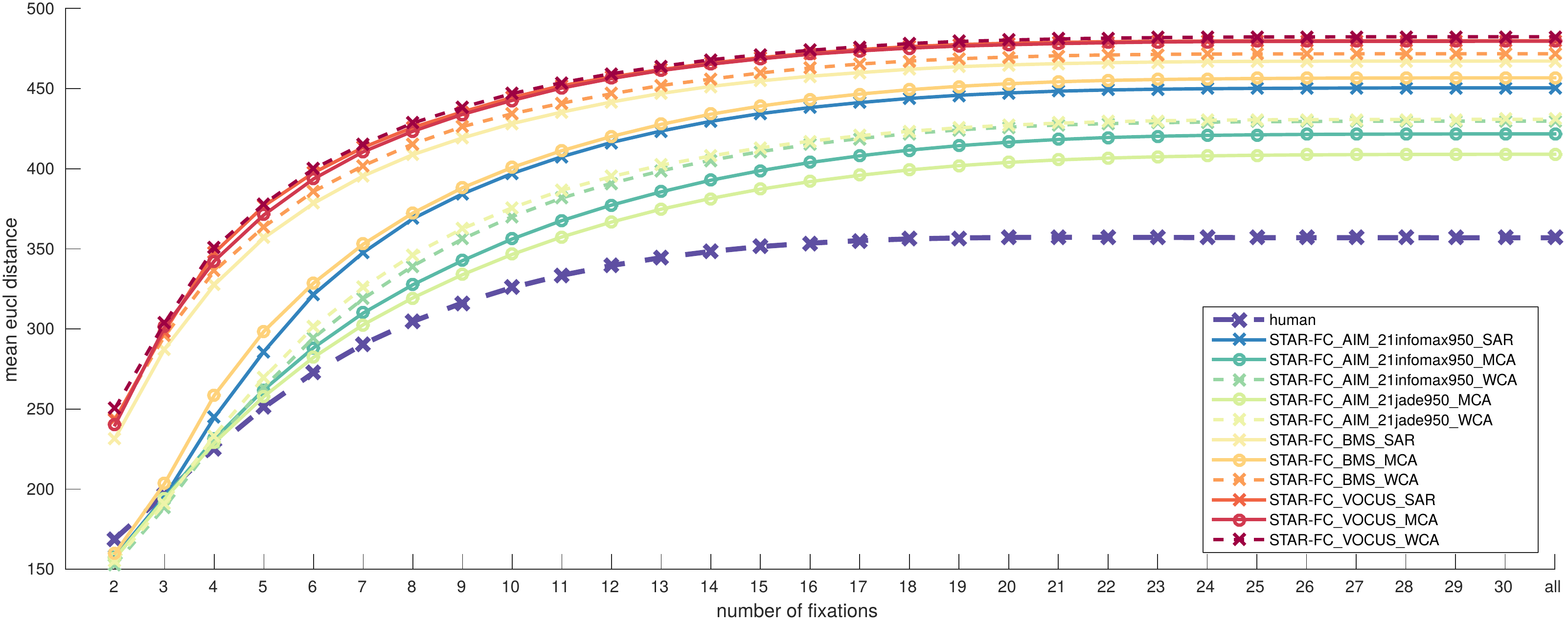}
			\caption{Euclidean distance}
		\end{subfigure}
		\begin{subfigure}[b]{0.75\textwidth}
			\includegraphics[width=\textwidth]{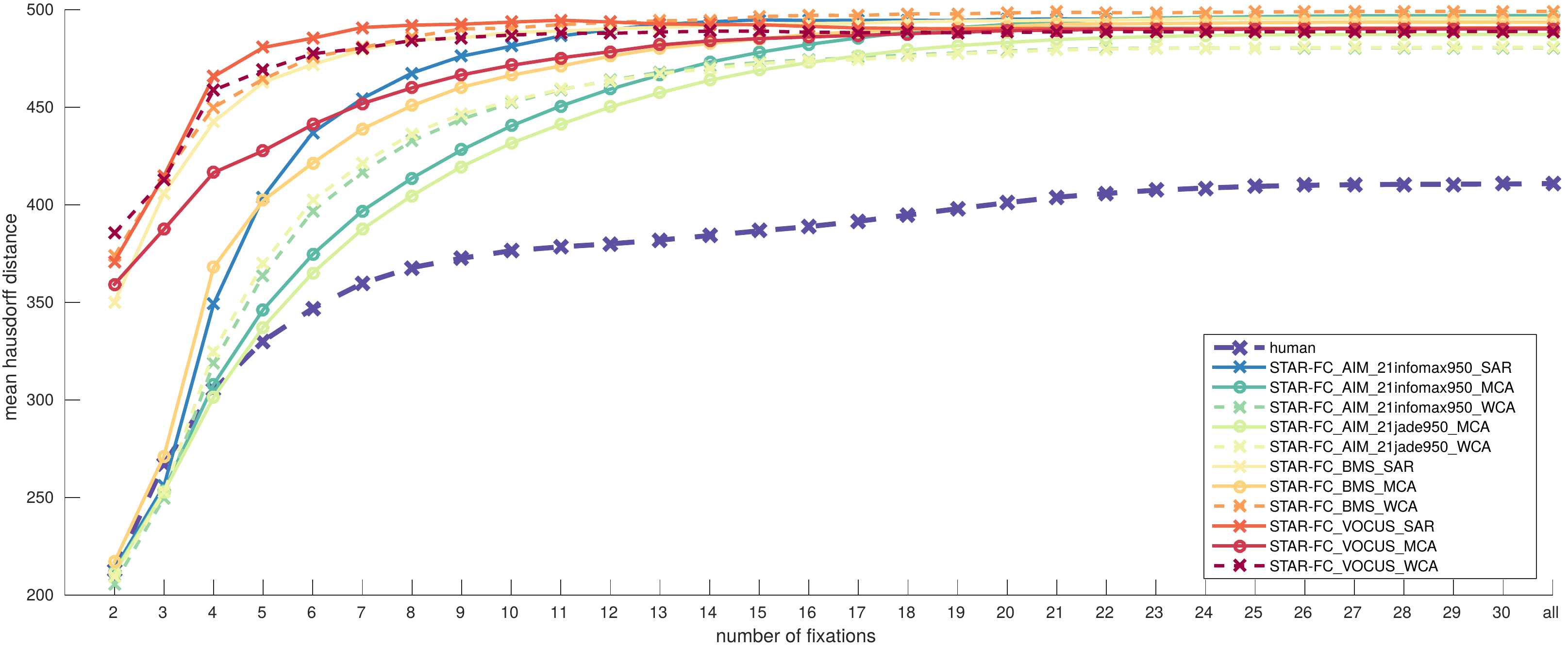}
			\caption{Hausdorff distance}
		\end{subfigure}
		\begin{subfigure}[b]{0.75\textwidth}
			\includegraphics[width=\textwidth]{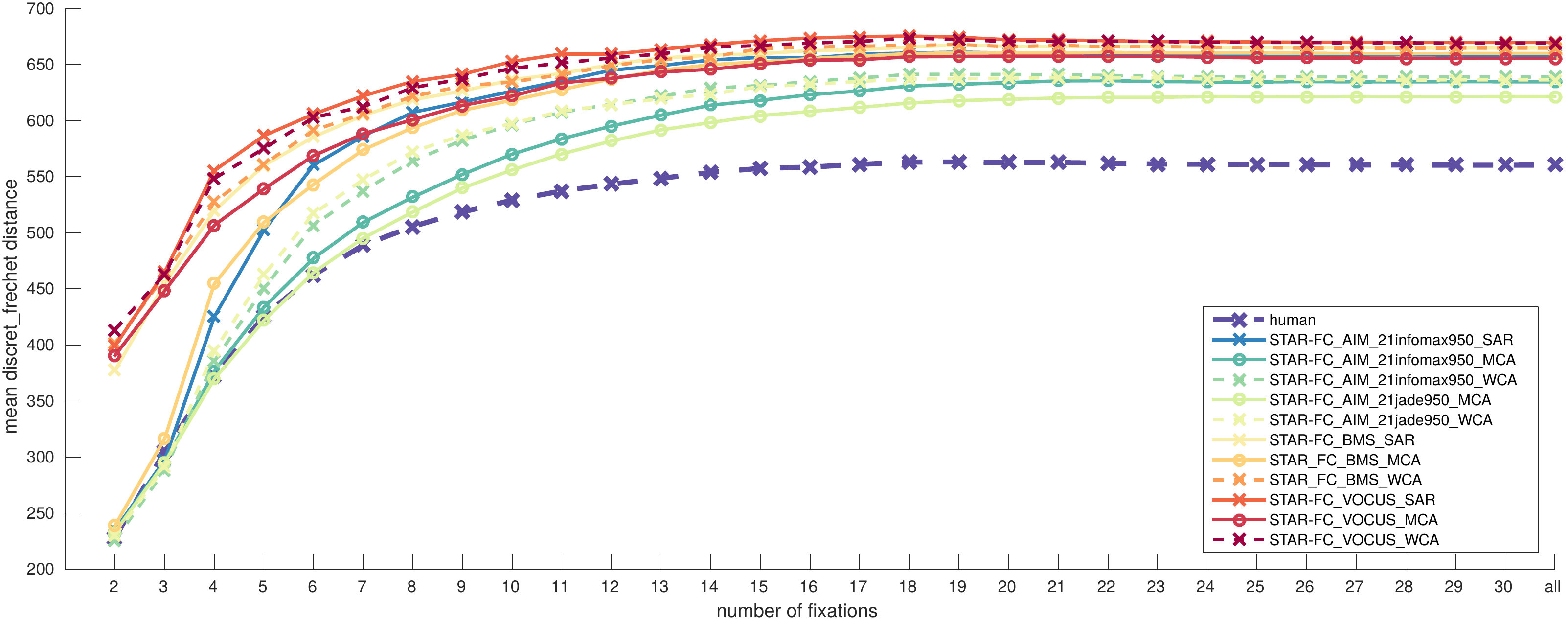}
			\caption{Frechet distance}
		\end{subfigure}
	\end{center}
	\vspace{-6mm}
	\caption{A comparison of fixation prediction scores over the full length of the fixation sequences for variants of STAR-FC.  
		\label{fig:FullTrajectoryScores}}
	\vspace{100mm}
\end{figure*}

\begin{figure*}[!htbp]
	\begin{center}
		\begin{subfigure}[b]{0.75\textwidth}
			\includegraphics[width=\textwidth]{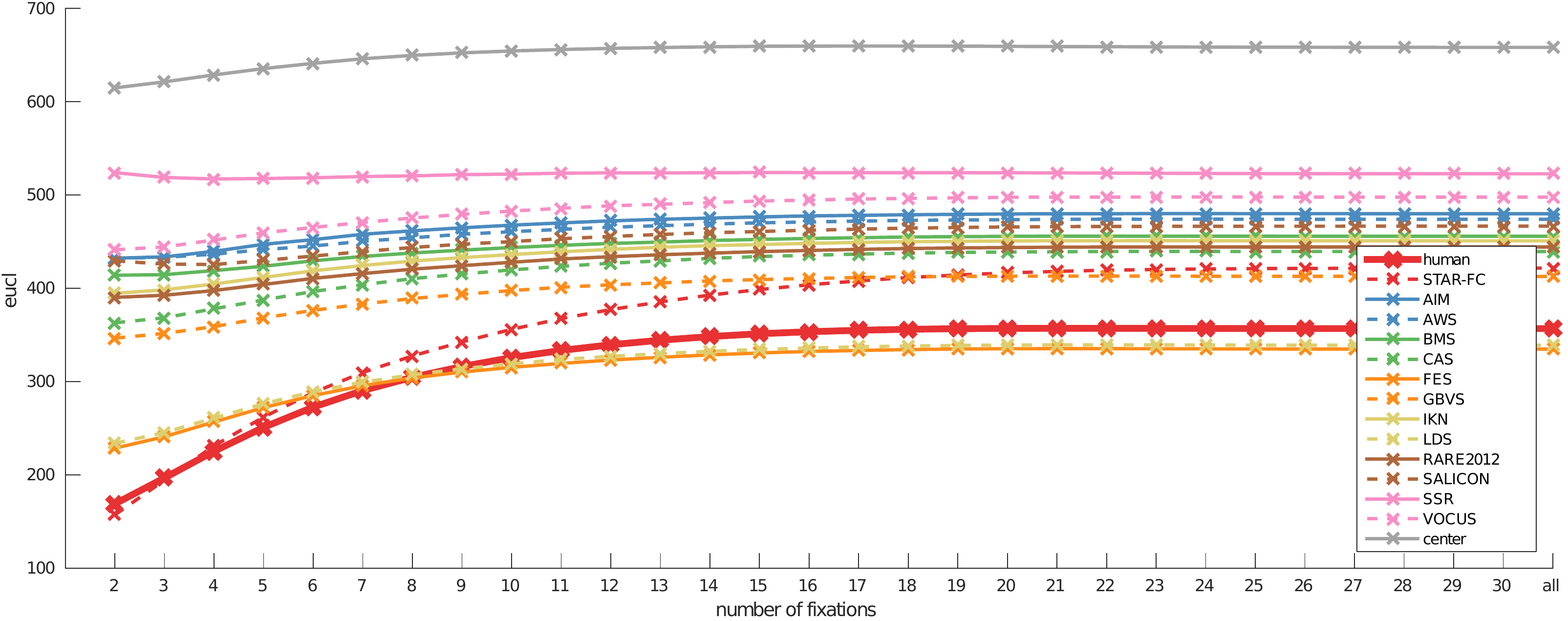}
			\caption{Euclidean distance}
		\end{subfigure}
		\begin{subfigure}[b]{0.75\textwidth}
			\includegraphics[width=\textwidth]{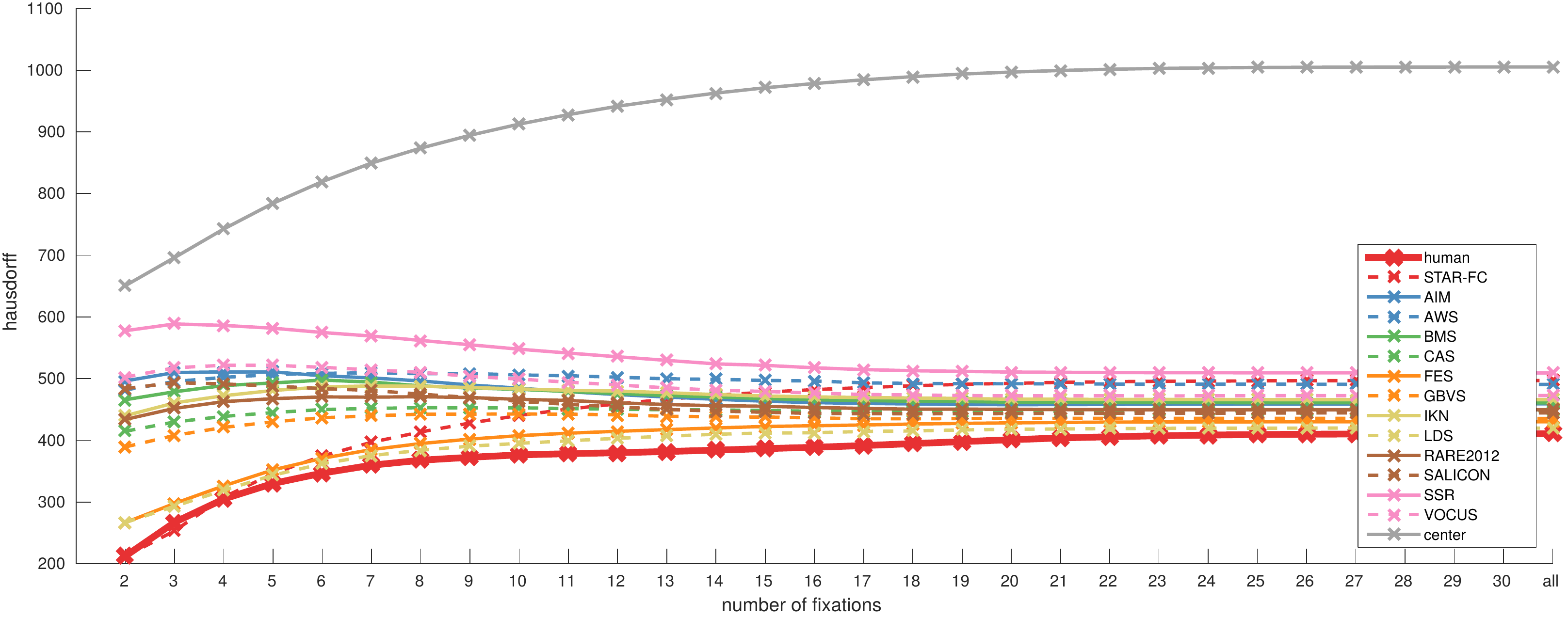}
			\caption{Hausdorff distance}
		\end{subfigure}
		\begin{subfigure}[b]{0.75\textwidth}
			\includegraphics[width=\textwidth]{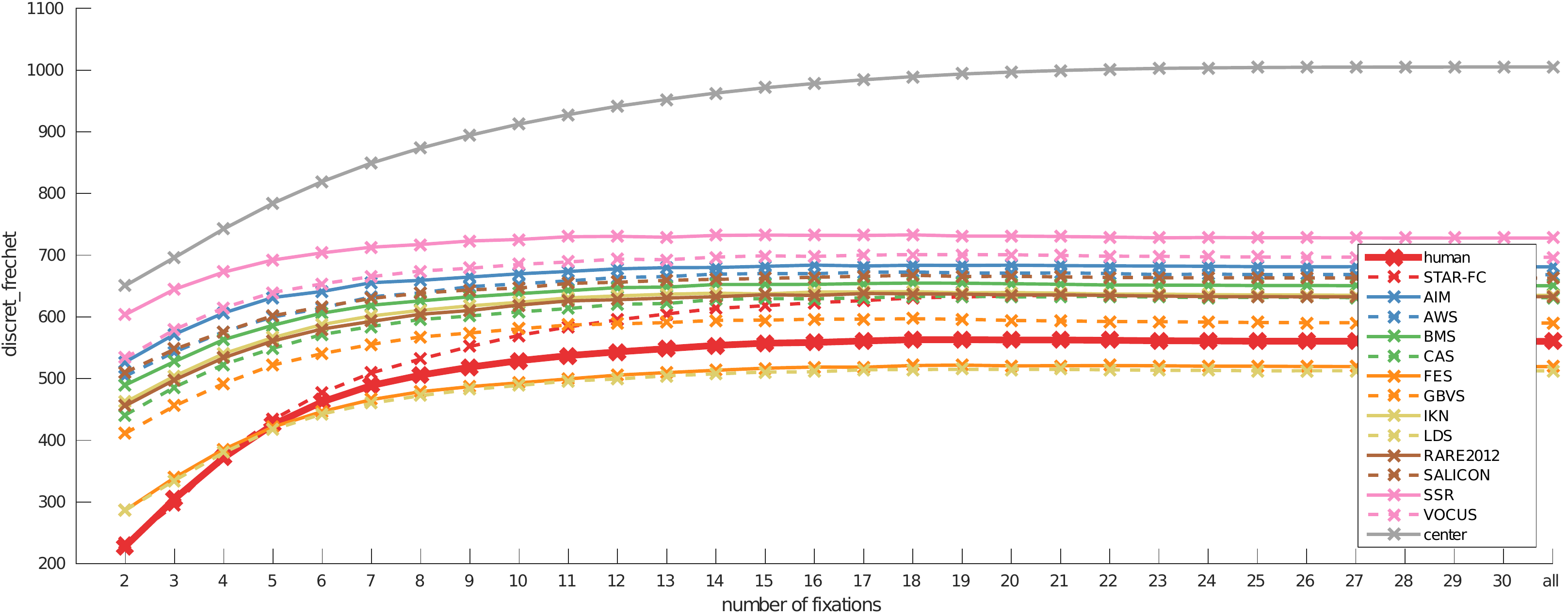}
			\caption{Frechet distance}
		\end{subfigure}
	\end{center}
	\vspace{-6mm}
	\caption{A comparison of fixation prediction scores over the full length of the fixation sequences for all tested saliency algorithms and the best performing STAR-FC model (using AIM with 21infomax950 basis in the peripheral field and MCA blending strategy). 
		\label{fig:SalFullTrajectoryScores}}
	\vspace{100mm}
\end{figure*}

\figref{fig:FullTrajectoryScores} and \figref{fig:SalFullTrajectoryScores} show trajectory scores for full sequence length for various STAR-FC variants and all tested saliencey algorithms. Note that as the sequences get longer they begin to diverge and the trajectory scores saturate.  

\begin{figure*}[!htbp]
	\begin{center}
		\begin{subfigure}[b]{0.75\textwidth}
			\includegraphics[width=\textwidth]{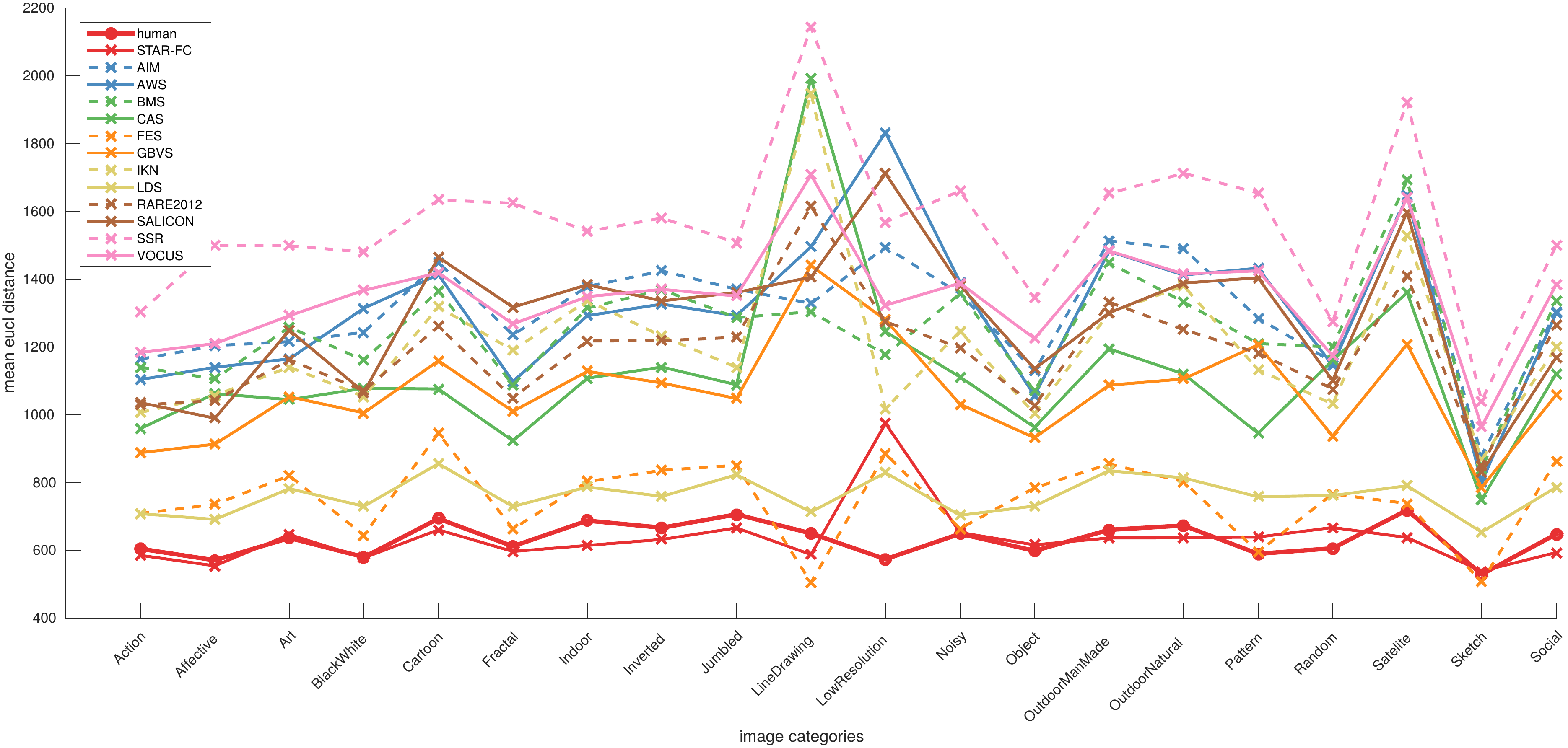}
			\caption{Euclidean distance}
		\end{subfigure}
		\begin{subfigure}[b]{0.75\textwidth}
			\includegraphics[width=\textwidth]{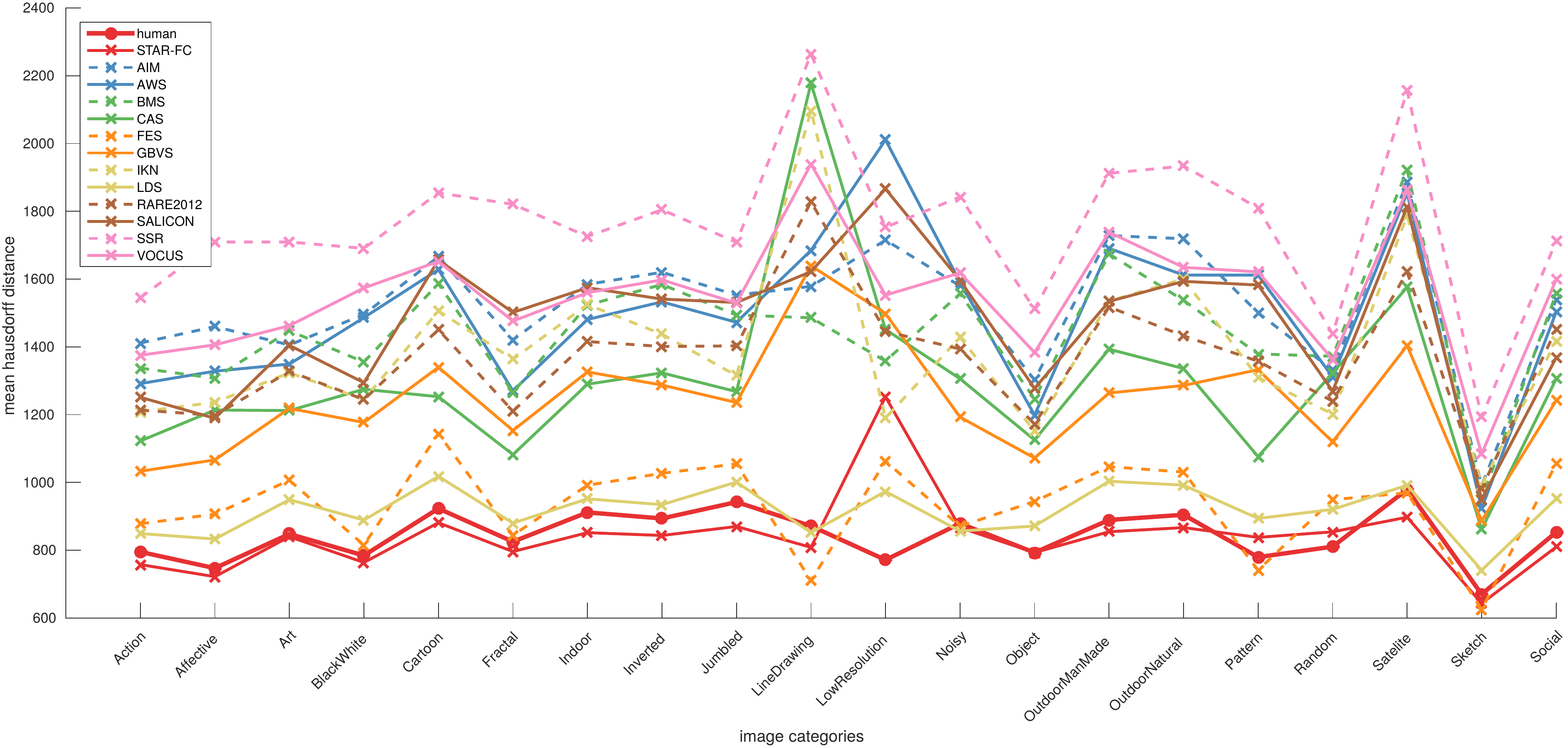}
			\caption{Hausdorff distance}
		\end{subfigure}
		\begin{subfigure}[b]{0.75\textwidth}
			\includegraphics[width=\textwidth]{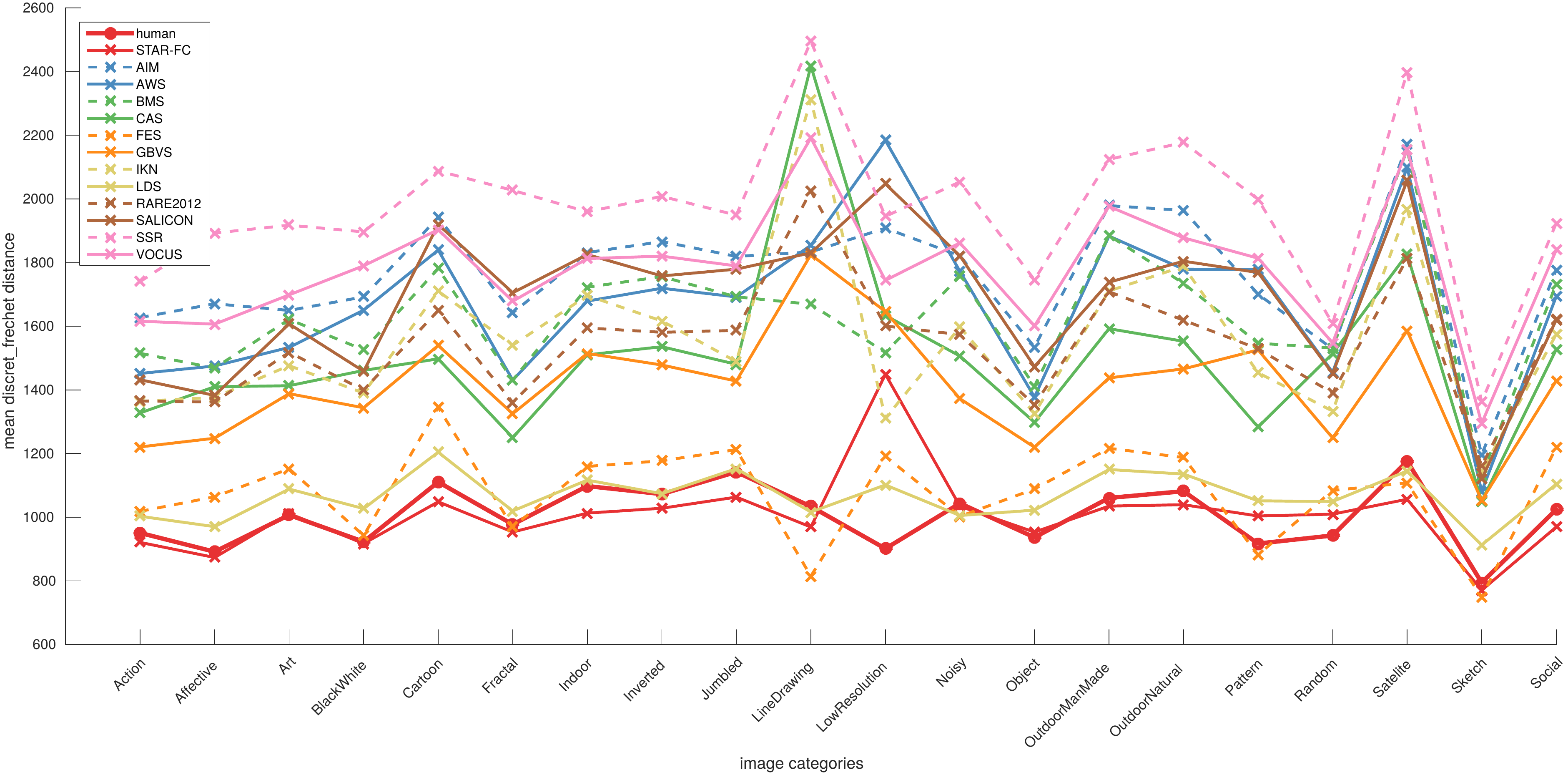}
			\caption{Frechet distance}
		\end{subfigure}
	\end{center}
	\vspace{-6mm}
	\caption{Fixation prediction scores for all tested saliency algorithms and the best performing STAR-FC model (using AIM with 21infomax950 basis in the peripheral field and MCA blending strategy). For each category we measured the mean distance from the human fixation and plotted the area-under-the-curve (AUC) score for the first 5 fixations.  
		\label{fig:SaliencyMetricsPerCategory}}
	\vspace{4mm}
\end{figure*}

\figref{fig:SaliencyMetricsPerCategory} shows the AUC score for the first 5 fixations for all saliency algorithms and our best performing STAR-FC model (using AIM with 21infomax950 basis and MCA blending strategy) split by category. For human fixations we report the AUC for the average pairwise distance. As we noted in the paper, it correlates well with inter-observer consistency for different categories of images (e.g. high IO consistency for Sketch translates to a smaller average pairwise distances across all metrics, whereas the opposite is true for the Jumbled category). 

Note that saliency algorithms tend to follow the same trends as human inter-observer scores. In general, categories with high IO consistency such as Affective or Sketch are not as challenging as categories with lower human to human fixation consistency. One major exception is the Low Resolution category, which has a high degree of IO consistency but is nevertheless extremely challenging for all saliency algorithms. This calls for more investigation of the effects that blurring has on the quality of saliency prediction. 

\section{Examples of Predicted Fixation Sequences}
\label{sec:ExamplesApdx}
Below we show some examples of predicted fixations. For clarity we only compare STAR-FC and one saliency algorithm at a time and show only the closest human sequences to each of the predicted sequences. Furthermore, we show results only for the first 3 and 5 fixations. We selected FES as the top performing contrast-based algorithm and SALICON as the top performing CNN-based algorithm for comparison with our best performing STAR-FC model (21infomax950 bases and MCA blending strategy).

\begin{figure*}[!htbp]
	\begin{center}
		\begin{subfigure}[b]{0.49\textwidth}
			\includegraphics[width=\textwidth]{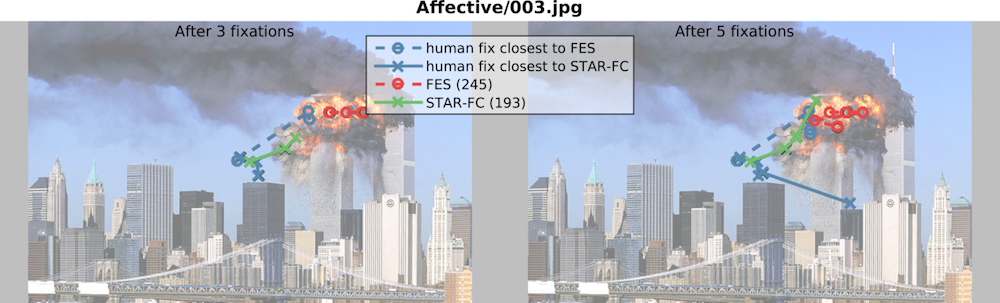}
		\end{subfigure}
		\begin{subfigure}[b]{0.49\textwidth}
			\includegraphics[width=\textwidth]{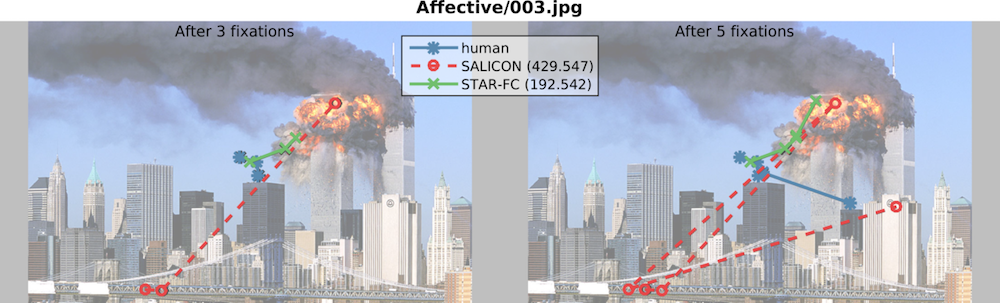}
		\end{subfigure}
		\begin{subfigure}[b]{0.49\textwidth}
			\includegraphics[width=\textwidth]{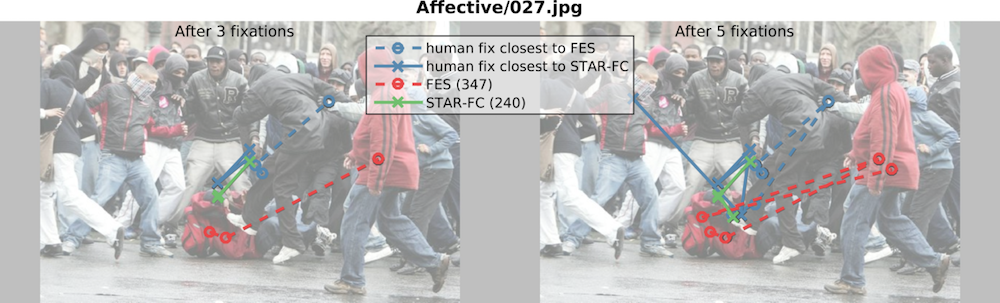}
		\end{subfigure}
		\begin{subfigure}[b]{0.49\textwidth}
			\includegraphics[width=\textwidth]{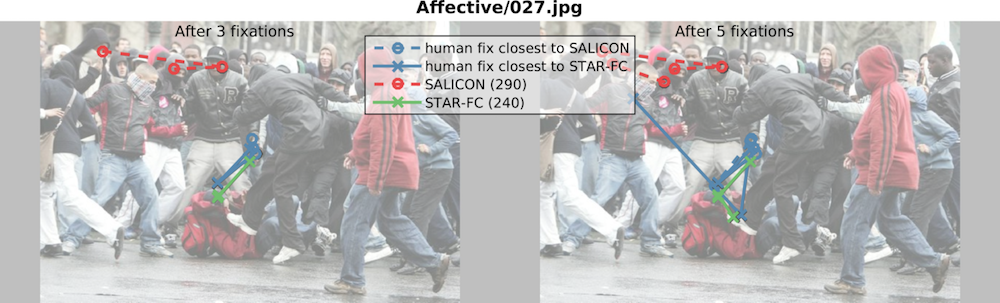}
		\end{subfigure}
		\begin{subfigure}[b]{0.49\textwidth}
			\includegraphics[width=\textwidth]{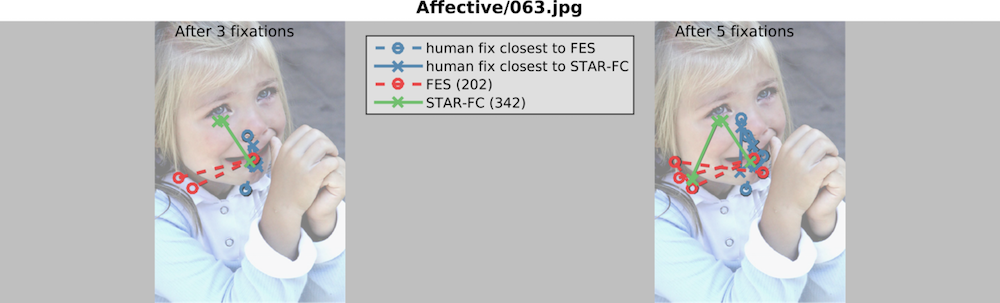}
		\end{subfigure}
		\begin{subfigure}[b]{0.49\textwidth}
			\includegraphics[width=\textwidth]{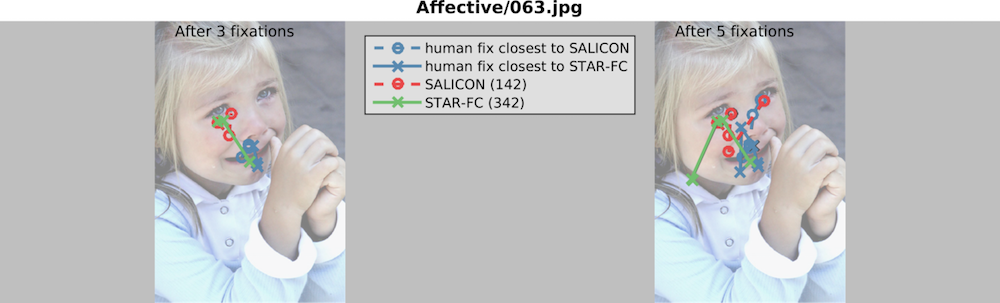}
		\end{subfigure}
		\begin{subfigure}[b]{0.49\textwidth}
			\includegraphics[width=\textwidth]{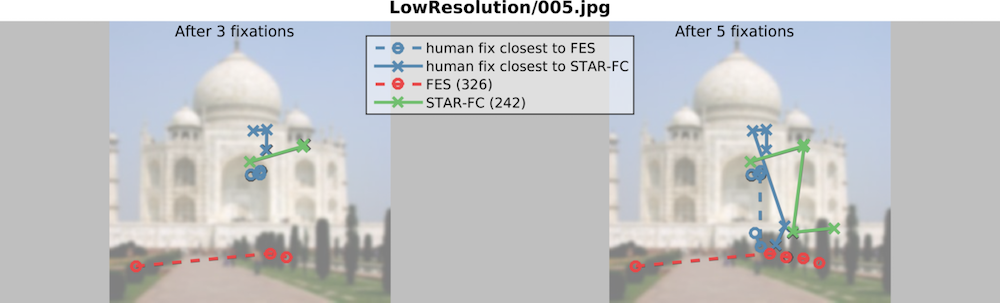}
		\end{subfigure}
		\begin{subfigure}[b]{0.49\textwidth}
			\includegraphics[width=\textwidth]{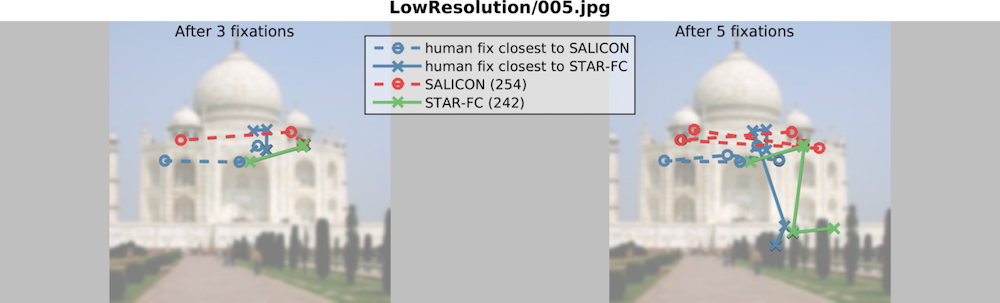}
		\end{subfigure}
		\begin{subfigure}[b]{0.49\textwidth}
			\includegraphics[width=\textwidth]{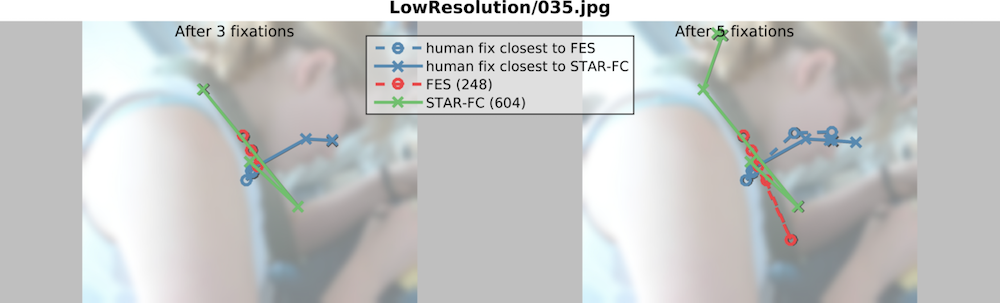}
		\end{subfigure}
		\begin{subfigure}[b]{0.49\textwidth}
			\includegraphics[width=\textwidth]{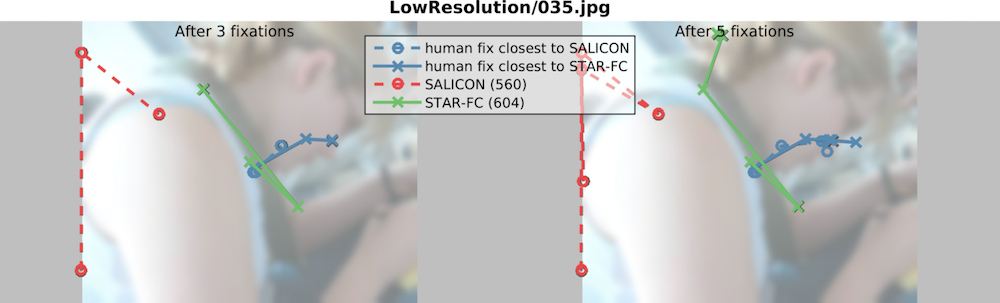}
		\end{subfigure}

\end{center}
\caption{Examples of fixations predicted by FES (left column) and SALICON (right column) compared to the proposed STAR-FC model (with AIM 21infomax basis and MCA blending strategy). Blue lines represent the closest human fixations to each of the compared algorithms and numbers in parentheses indicate the corresponding Euclidean distance.\label{fig:FixationExamplesAffectiveLowRes}}
\end{figure*}

\begin{figure*}[!htbp]
	\begin{center}
		\begin{subfigure}[b]{0.49\textwidth}
			\includegraphics[width=\textwidth]{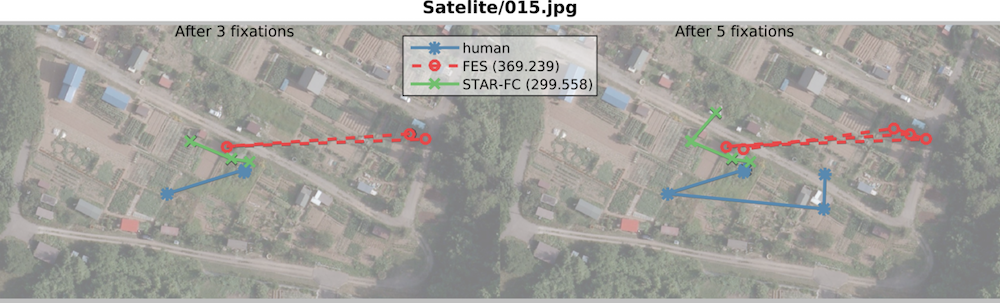}
		\end{subfigure}
		\begin{subfigure}[b]{0.49\textwidth}
			\includegraphics[width=\textwidth]{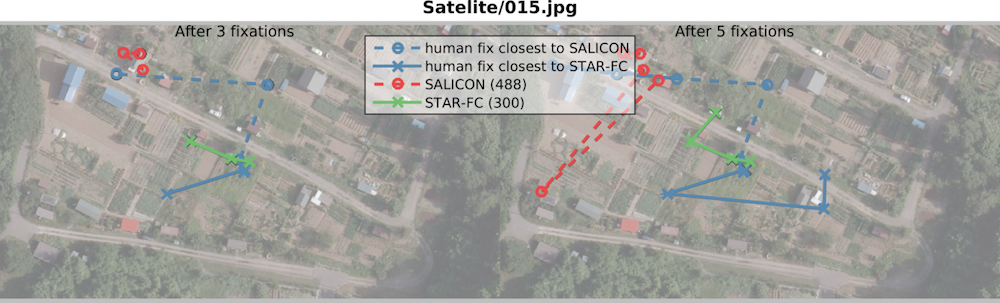}
		\end{subfigure}
		\begin{subfigure}[b]{0.49\textwidth}
			\includegraphics[width=\textwidth]{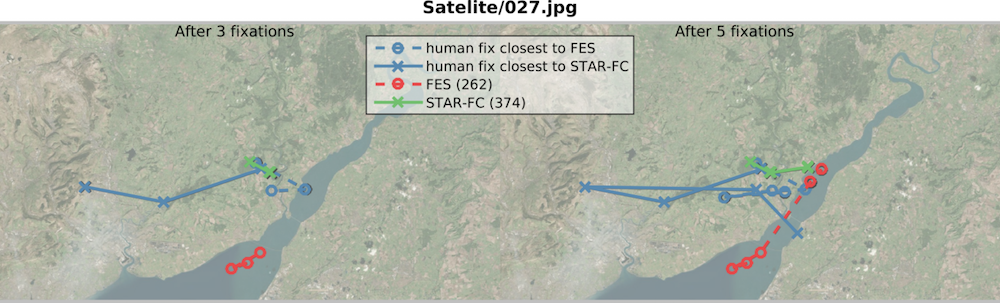}
		\end{subfigure}
		\begin{subfigure}[b]{0.49\textwidth}
			\includegraphics[width=\textwidth]{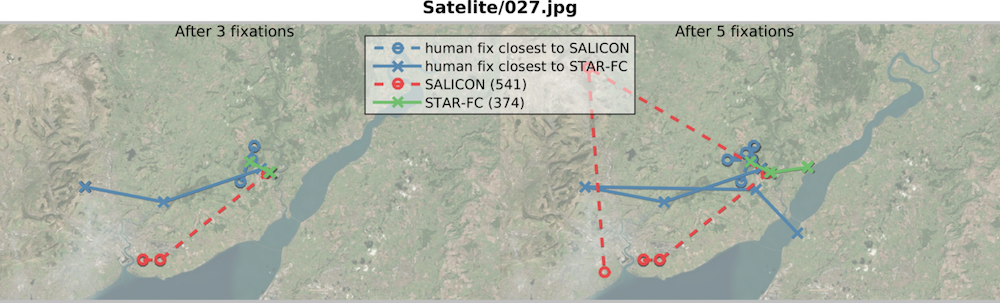}
		\end{subfigure}
		\begin{subfigure}[b]{0.49\textwidth}
			\includegraphics[width=\textwidth]{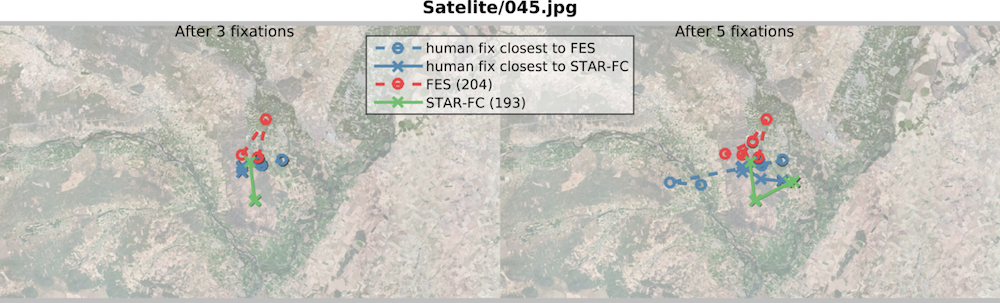}
		\end{subfigure}
		\begin{subfigure}[b]{0.49\textwidth}
			\includegraphics[width=\textwidth]{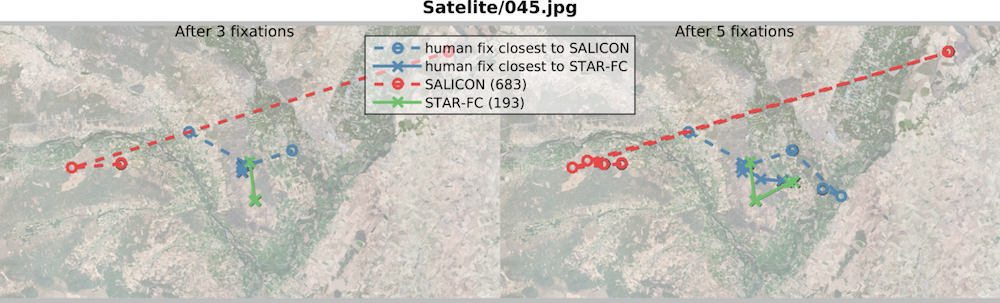}
		\end{subfigure}
\end{center}
\caption{Examples of fixations predicted by FES (left column) and SALICON (right column) compared to the proposed STAR-FC model (with AIM 21infomax basis and MCA blending strategy). Blue lines represent the closest human fixations to each of the compared algorithms and numbers in parentheses indicate the corresponding Euclidean distance.\label{fig:FixationExamplesSatelite}}
\end{figure*}

In \figref{fig:FixationExamplesAffectiveLowRes} examples from categories with high IO consistency (Affective and Low Resolution) are shown. \figref{fig:FixationExamplesSatelite} shows examples from the Satelite category which has low IO consistency.

\end{appendices}

\end{document}